\documentclass[preprint,12pt,numbers]{elsarticle}

\usepackage{amsmath,amssymb,amsfonts}
\usepackage{amsthm}
\usepackage{stmaryrd} 
\usepackage{graphicx}
\setkeys{Gin}{width=\linewidth,keepaspectratio}
\usepackage{textcomp}
\usepackage{booktabs}
\usepackage{tabularx} 
\usepackage{xcolor}
\usepackage{microtype} 
\usepackage{fancyvrb} 
\usepackage{fvextra}  

\usepackage[hidelinks]{hyperref}
\usepackage{enumitem} 
\usepackage{pifont} 
\usepackage{algorithm}
\usepackage{algpseudocode}

\usepackage{tikz}
\usetikzlibrary{positioning,calc,arrows.meta,shapes.geometric}

\newcommand{\cmark}{\ding{51}}
\newcommand{\xmark}{\ding{55}}

\theoremstyle{definition}
\newtheorem{definition}{Definition}

\newcommand{\multiset}[1]{\{\!\{#1\}\!\}}

\graphicspath{{figures/}}

\IfFileExists{generated_metrics.tex}{

}{%
}

\begin{document}

\begin{frontmatter}

\title{Schema-Aware Localisation (SAL): Live Schema Grounding and Hallucination Validation for Oracle NL2SQL}

\author[inst1]{Sanjay Mishra\fnref{fn:smieee}}
\ead{sanmish4@icloud.com}

\author[inst2]{Divya Chukkapalli\fnref{fn:smieee}}
\ead{divya.95j@gmail.com}

\author[inst3]{Ganesh R. Naik\fnref{fn:smieee}}
\ead{ganesh.naik@torrens.edu.au}

\fntext[fn:smieee]{Senior Member, IEEE.}

\address[inst1]{Independent Researcher, Raleigh, NC 27601, USA}
\address[inst2]{Independent Researcher, Apex, NC 27502, USA}
\address[inst3]{Torrens University Australia, Adelaide, SA, Australia}

\begin{abstract}
Large language models can generate fluent SQL from natural language, but on
real enterprise Oracle databases they frequently fail at execution time:
columns and aliases are hallucinated and dialect-specific syntax is missed,
leading to \texttt{ORA-00904} invalid-identifier errors. In this setting,
failures are primarily due to missing schema grounding: the model cannot
know which tables and columns actually exist.

This paper introduces \textbf{Schema-Aware Localisation (SAL)}, a lightweight
middleware layer for Oracle NL2SQL that requires no model retraining. SAL
queries Oracle's \texttt{USER\_TAB\_COLUMNS} catalog to build a live schema map,
selects a relevant table subset for each question (falling back to the full
schema for multi-table queries), and injects this ground-truth context into
the LLM prompt. Generated SQL is then checked by the Hallucination Index
(Hidx), which validates every \texttt{alias.column} reference against the live
catalog, automatically rewrites predictable prefix errors, and otherwise
triggers a structured retry with itemised corrections.

We evaluate SAL on 500 TPC-H natural language questions executed against a
live Oracle Autonomous Database~23c instance using GPT-4o-mini. Without any
schema grounding, execution-grounded truth (EGT; executes and matches the
reference result set) is 2.2\% (12/500). A hand-written static schema hint
brings EGT to 62.0\%. SAL, with no manual schema curation, achieves
\textbf{62.6\% EGT} (96\% simple, 95\% medium, 40.7\% complex) while reducing
execution failures from 97.6\% to 2.6\%.

\end{abstract}


\end{frontmatter}

\section{Introduction}

Databases do not speak English.  For decades, this has been the quiet tax
on enterprise analytics: a business analyst with a sharp question must wait
for a database engineer to translate it into SQL, or learn the query language
themselves.  Large language models (LLMs) appeared to change this.  Systems
like GPT-4o-mini can produce syntactically fluent SQL from plain English
questions in under a second, and on curated benchmarks such as
Spider~\cite{Yu2018} they achieve accuracy figures above 80\%.

The gap between benchmark performance and production reality, however, is
stark.  Spider queries run against clean, well-documented schemas with
standard column names.  Enterprise Oracle databases do not look like
Spider.  They carry decades of naming conventions: columns prefixed with
table initials (\texttt{O\_ORDERDATE}, \texttt{L\_SHIPDATE},
\texttt{C\_CUSTKEY}), constraints enforced at the dialect level
(\texttt{FETCH FIRST} instead of \texttt{LIMIT}, no
\texttt{EXTRACT(QUARTER)}), and internal catalog structures that no LLM has
seen during training.  When a model is asked to query such a database without
being told what columns exist, it guesses, and Oracle does not forgive
guesses.  The result is a wall of \texttt{ORA-00904: invalid identifier}
errors.

Our own measurements confirm how severe this problem is.  In a controlled
experiment over 500 natural language questions on a live Oracle Autonomous
Database instance, GPT-4o-mini with no schema context achieved just
\textbf{2.2\%} execution-grounded truth (EGT), our primary accuracy metric:
the fraction of questions for which the generated SQL both executes and
returns the correct result on the live database.  Only 12 of 500 queries
executed and returned correct results.  The remaining 488 failed at the
Oracle execution layer, not because the model reasoned poorly, but because
it had no way to know which columns existed.

This is primarily a grounding problem in this setting, and it has a
practical solution that does not require retraining.  Oracle exposes schema
metadata through data dictionary views such as \texttt{USER\_TAB\_COLUMNS}.
If that metadata is fetched at runtime and injected into the LLM prompt, the
model gains the ground truth it was missing.  A hand-crafted static hint
built from this metadata raises \textbf{EGT to 62.0\%} in our experiments,
a gain of nearly 60 percentage points from a few lines of prompt engineering.

The challenge, then, is automating and hardening this grounding so that it
works across heterogeneous Oracle deployments (with evolving naming
conventions, privileges, and dialect constraints) and across the question
types encountered in practice, without a human writing and maintaining the
hint.  This is the problem SAL addresses.

SAL differs from generic ``schema-in-the-prompt'' approaches in three ways.
First, it targets Oracle-specific production failure modes (dialect features
and identifier conventions) and grounds against the live data dictionary at
runtime rather than a manually curated schema string.  Second, it couples
selective schema retrieval with a complexity-gated fallback to avoid
under-scoping multi-table questions.  Third, it enforces schema validity with
an execution-facing validator (Hidx) that can rewrite predictable identifier
prefix errors and otherwise drives structured retries with itemised
corrections.

\subsection*{Contributions}

This paper presents \textbf{Schema-Aware Localisation (SAL)}, a
zero-retraining runtime grounding framework for Oracle NL2SQL, implemented
as an open-source Model Context Protocol (MCP)~\cite{Anthropic2024} server.
The specific contributions are:

\begin{enumerate}

  \item \textbf{Live schema injection.}  SAL queries \texttt{USER\_TAB\_COLUMNS}
  at server startup and maintains an in-memory cache mapping every table to
  its exact ordered column list.  This cache is refreshed per deployment and
  requires no manual maintenance.

  \item \textbf{Complexity-gated table detection (SAL Detect v2).}
  Rather than binary keyword matching, SAL v2 scores each table by the number
  of keyword hits in the question, expands the matched set along known
  JOIN-chain edges (e.g., matching \textsc{Orders} also includes
  \textsc{Customer} and \textsc{Lineitem}), and falls back to the full schema
  when the question contains multi-table complexity signals and fewer than
  three tables match directly.  This prevents the ``narrow hint'' failure
  mode in which JOIN-partner tables are silently dropped.

  \item \textbf{Hallucination Index (Hidx) validator.}  To reduce Oracle
  execution failures and enforce schema-valid SQL under live schemas, Hidx
  parses each generated statement, builds an alias-to-table map from
  \texttt{FROM}/\texttt{JOIN} clauses, and checks every \texttt{alias.column}
  reference against the live schema cache.  References that follow a
  predictable prefix pattern (e.g., \texttt{O.ORDERDATE} instead of
  \texttt{O.O\_ORDERDATE}) are rewritten automatically.  Others trigger a
  structured retry in which the LLM receives an itemised list of its invalid
  references and the correct alternatives.

  \item \textbf{Empirical ablation over Oracle ADB.}  We evaluate four
  conditions (no hint, static hint, SAL v1, and SAL v2) on 500 TPC-H
  questions across three complexity tiers, executed against a live Oracle~23c
  instance.  SAL v2 achieves \textbf{62.6\% EGT} (execution-grounded truth;
  correct result on the live database), matching the hand-crafted static
  baseline and reducing execution failures from 97.6\% to 2.6\%, with no
  manual schema work.

\end{enumerate}

\section{Related Work}
\label{sec:related}

The literature relevant to this work spans four areas: natural language
interfaces to databases, schema linking, hallucination detection in
generative models, and tool-augmented language model architectures.  We
survey each in turn and position SAL with respect to prior art.

\subsection{Natural Language Interfaces to Databases}

The problem of translating natural language questions into structured
database queries has been studied since the 1970s, beginning with
rule-based systems such as LUNAR~\cite{Woods1973} and LADDER~\cite{Hendrix1978}.
Modern interest re-accelerated with the introduction of large-scale
annotated benchmarks.  WikiSQL~\cite{Zhong2017} provided 80,654
question--SQL pairs over single-table schemas and enabled systematic
comparison of neural approaches.  Spider~\cite{Yu2018} extended the
evaluation to 200 databases across 138 domains with complex, cross-table
queries requiring joins, subqueries, and aggregations.  BIRD~\cite{Li2023}
further raised the bar by introducing realistic database noise and
evidence-grounded questions.  These benchmarks collectively define common
evaluation conventions, including execution-based accuracy and string-level
exact-match.  In this work we focus on execution-grounded truth (EGT), i.e.,
the fraction of questions for which the generated SQL both executes and
returns the correct result on the live database, and we additionally report
a semantic-match count for diagnostic purposes.

Early deep learning approaches encoded questions and schemas with
sequence-to-sequence models~\cite{Zhong2017, Dong2016, Xu2017}, decoding
SQL tokens autoregressively.  TypeSQL~\cite{Yu2018b} incorporated type
information from schema metadata to improve column recognition.  ShadowGNN~\cite{Chen2021}
and IGSQL~\cite{Cai2020} extended these ideas to multi-turn conversational
settings where schema context must persist across dialogue turns.

The dominant paradigm shifted to large pre-trained transformers following
the success of BERT~\cite{Devlin2019} and T5~\cite{Raffel2020}.  BRIDGE~\cite{Lin2020}
serialised schema information directly into the input and used BERT
representations to align question tokens with database entities.
GRAPPA~\cite{Yu2021} pre-trained a language model specifically on SQL-grounded
corpora to improve schema understanding.  These approaches consistently
outperformed their predecessors on Spider but remain dependent on schema
representations fixed at training time.

With the emergence of instruction-tuned LLMs, few-shot prompting became the
dominant approach.  DAIL-SQL~\cite{Gao2023} selects question-similar
examples from a curated pool via masked question similarity and achieves
86.6\% execution accuracy on Spider using GPT-4.  DIN-SQL~\cite{Pourreza2023}
decomposes the problem into schema linking, query classification, query
generation, and self-correction sub-tasks, reaching 85.3\% on Spider with
GPT-4.  C3-SQL~\cite{Dong2023} improves robustness through consistency
voting across multiple independently generated candidates.

A limitation of many of the above systems in enterprise settings is that they
assume a clean, well-documented schema is readily available and stable at
inference time.  Far fewer works explicitly study the case where schema
metadata must be retrieved live from a proprietary database catalog and where
identifier conventions in the target database deviate sharply from the
training distribution.

\subsection{Schema Linking}

Schema linking, the sub-task of identifying which tables and columns a
natural language question refers to before SQL generation, is widely
recognised as the primary bottleneck in NL2SQL
accuracy~\cite{Wang2020, Lei2020, Cao2021}.  Errors in schema linking
propagate irreversibly into the generated query.

IRNet~\cite{Guo2019} introduced a concept vocabulary that maps question
spans to schema elements via string similarity and word-overlap heuristics.
RAT-SQL~\cite{Wang2020} replaced heuristic linking with a relation-aware
transformer that jointly encodes question tokens and schema entities in a
unified graph, learning link weights from annotated examples.
LGESQL~\cite{Cao2021} further refined this with line-graph-enhanced schema
encoding that captures both local and global schema structure.

These methods learn schema linking from thousands of annotated
question--schema pairs.  They generalise within the distribution of their
training schemas but degrade when column names follow enterprise conventions
not present in Spider or WikiSQL, such as the TPC-H prefix pattern
(\texttt{O\_ORDERDATE}, \texttt{L\_SHIPDATE}) or Oracle-internal naming of
system objects.

SAL takes an orthogonal approach: rather than learning to link question
tokens to a static schema vocabulary, it retrieves the canonical schema at
runtime from the database's own system catalog (\texttt{USER\_TAB\_COLUMNS})
and injects it as grounding context into the LLM prompt.  This requires no
annotated examples, no fine-tuning, and no advance knowledge of the target
schema, only a live database connection with read access to the relevant data
dictionary views.

\subsection{Hallucination Detection and Correction in Generative Models}

Hallucination, the generation of plausible but factually incorrect
content, is a well-documented failure mode of generative language
models~\cite{Maynez2020, Ji2023}.  In natural language generation,
hallucinations are difficult to detect automatically because they require
external fact verification.  In code and SQL generation, however,
hallucinations are directly and cheaply detectable: the runtime environment
(compiler, interpreter, or database engine) raises an error.

Liu et al.~\cite{Liu2023} evaluate LLM-generated code on 728 programming
problems and find that GPT-4 produces incorrect code in 37\% of cases
despite generating syntactically valid programs. For SQL specifically,
Ni et al.~\cite{Ni2023} study execution-guided decoding strategies that
use database feedback to constrain SQL generation, reducing execution
errors. Pourreza and Rafiei~\cite{Pourreza2023} incorporate a
self-correction step in DIN-SQL where the LLM is shown its own Oracle
error output and asked to revise; this reduces failures but does not
identify the specific erroneous token.

Our Hallucination Index (Hidx) advances beyond execution-guided and
self-correction approaches in three ways.  First, it operates
\emph{pre-execution}: errors are detected by static analysis of the
generated SQL against the live schema cache, avoiding an unnecessary round
trip to the database.  Second, it is \emph{token-precise}: Hidx identifies
the exact \texttt{alias.column} reference that is invalid and the table it
is mapped to, rather than relying on a generic runtime error message.
Third, it distinguishes \emph{auto-correctable} errors (those matching a
predictable alias-prefix pattern such as \texttt{O.ORDERDATE} for
\texttt{O.O\_ORDERDATE}) from errors that require LLM re-prompting, saving
API calls for the former class.

\subsection{Oracle SQL and Enterprise Database Deployment}

The overwhelming majority of NL2SQL research targets open-source SQL
dialects (SQLite, PostgreSQL, and MySQL) under controlled benchmark
conditions.  Oracle SQL introduces a qualitatively different set of
challenges.

Oracle enforces strict identifier scoping: a column referenced as
\texttt{alias.COLUMN} must match the column identifier as stored in
\texttt{USER\_TAB\_COLUMNS} (typically uppercased unless quoted).  The TPC-H
schema, in its Oracle instantiation, uses table-prefixed column names
(\texttt{O\_ORDERKEY}, \texttt{L\_SHIPDATE}, \texttt{C\_CUSTKEY}) that do
not appear in any NL2SQL training corpus.  Common LLM outputs such as
\texttt{O.ORDERDATE} or \texttt{O.ORDER\_DATE} produce
\texttt{ORA-00904: invalid identifier} at runtime regardless of semantic
correctness.  Oracle also prohibits \texttt{LIMIT} (requiring
\texttt{FETCH FIRST n ROWS ONLY}), lacks \texttt{EXTRACT(QUARTER)}, and
handles analytic window functions under different syntactic constraints than
PostgreSQL~\cite{Oracle2024}.

We are not aware of a public NL2SQL benchmark or ablation study that makes
Oracle dialect correctness and live-catalog grounding first-class measurement
dimensions on a live Oracle Autonomous Database instance.  The closest
related work is on enterprise SQL assistance tools such as Oracle APEX
AI~\cite{OracleAPEX2024} and commercial text-to-SQL products, which provide
LLM-powered query generation but generally do not publish rigorous accuracy
ablations against live Oracle instances or quantify the contribution of schema
grounding to execution correctness.

\subsection{Tool-Augmented Language Models and Model Context Protocol}

Tool-augmented LLMs, systems that invoke external APIs, retrieve documents,
or execute programs during inference, have demonstrated that grounding LLM
outputs in retrieved evidence substantially reduces hallucination and improves
factual accuracy~\cite{Schick2023, Yao2023, Lewis2020}.  Retrieval-Augmented
Generation (RAG)~\cite{Lewis2020} retrieves relevant passages from a document
corpus and prepends them to the LLM input; the retrieved text serves as an
authoritative reference that constrains generation.  SAL applies this
principle specifically to the database schema domain, where the retrieved
evidence (column metadata from \texttt{USER\_TAB\_COLUMNS}) is uniquely
authoritative, machine-readable, and verifiable, enabling automated
post-hoc validation that document-based RAG cannot support.

The Model Context Protocol (MCP)~\cite{Anthropic2024} is an emerging open
standard that defines a structured HTTP interface for exposing tools and data
sources to LLMs.  MCP provides a uniform integration layer analogous to what
USB provides for peripheral devices.  SAL is implemented as an MCP server,
enabling it to be deployed alongside any MCP-compatible LLM client, including
IDE assistants, chatbots, or business intelligence tools, without
application-level code changes.  This
separates the grounding and validation logic from the consumer application
and allows a single SAL instance to serve multiple clients concurrently.

\subsection{Research Gap and Positioning}

Table~\ref{tab:related} summarises the key dimensions on which SAL differs
from prior work.

\begin{table}[t]
  \centering
  \small
  \setlength{\tabcolsep}{4pt}
  \renewcommand{\arraystretch}{1.2}
  \caption{Comparison of NL2SQL systems on enterprise Oracle deployment dimensions.}
  \label{tab:related}
  \resizebox{\columnwidth}{!}{%
  \begin{tabular}{lccccc}
    \toprule
    \textbf{System} & \textbf{\shortstack{Live\\Schema}} &
                    \textbf{\shortstack{Halluc.\\Validator}} &
                    \textbf{\shortstack{Oracle\\Dialect}} &
                    \textbf{\shortstack{No\\Retrain}} &
                    \textbf{\shortstack{MCP\\Ready}} \\
    \midrule
    RAT-SQL~\cite{Wang2020}     & \xmark & \xmark              & \xmark & \xmark & \xmark \\
    DAIL-SQL~\cite{Gao2023}     & \xmark & \xmark              & \xmark & \cmark & \xmark \\
    DIN-SQL~\cite{Pourreza2023} & \xmark & \textit{Partial}     & \xmark & \cmark & \xmark \\
    \textbf{SAL (this work)}   & \cmark & \cmark              & \cmark & \cmark & \cmark \\
    \bottomrule
  \end{tabular}}
\end{table}

The research gap is specific: no existing NL2SQL system combines (i) live
catalog-driven schema injection, (ii) pre-execution column-reference
validation, (iii) Oracle dialect awareness, and (iv) zero-retraining
deployment in a single framework.  Commercial tools address (iv) but not
(i)--(iii); academic systems address (iv) but not (i)--(iii); and Oracle-specific
tooling exists but publishes no rigorous accuracy evaluation.  SAL is,
to our knowledge, the first system to address all four simultaneously, and
this paper provides the first published ablation quantifying the accuracy
contribution of each component on a live Oracle Autonomous Database instance.

\section{Background and Definitions}
\label{sec:background}

This section establishes the formal definitions and background concepts
referenced throughout the paper. Readers familiar with NL2SQL evaluation
and Oracle database administration may proceed directly to
Section~\ref{sec:design}.


\subsection{Oracle SQL Dialect}

Oracle Database enforces a number of syntactic and semantic constraints that
differ from the ANSI SQL standard and from the PostgreSQL and SQLite dialects
dominant in NL2SQL benchmarks.  The following are particularly relevant to
this work.

\begin{definition}[Oracle Row Limiting]
Oracle does not support the \texttt{LIMIT} clause.  Row limiting is expressed
using the \texttt{FETCH FIRST} syntax introduced in Oracle 12c:
\begin{verbatim}
SELECT ... FROM ... ORDER BY ...
FETCH FIRST n ROWS ONLY;
\end{verbatim}
Alternatively, the legacy \texttt{ROWNUM} pseudo-column may be used in a
subquery filter, but this does not compose correctly with \texttt{ORDER BY}.
\end{definition}

\begin{definition}[Oracle Column Identifier Scoping]
In Oracle, a column reference of the form \texttt{alias.column} is resolved
by an exact, case-insensitive match against the column names stored in
\texttt{USER\_TAB\_COLUMNS.COLUMN\_NAME}.  If \texttt{COLUMN} does not
appear in that view for the table bound to \texttt{alias}, Oracle raises
\texttt{ORA-00904: "alias"."COLUMN": invalid identifier}.  There is no
fuzzy matching or partial-name resolution.
\end{definition}

\begin{definition}[TPC-H Column Prefix Convention]
The TPC-H benchmark schema~\cite{TPC2022} names all columns with a
two-letter table prefix followed by an underscore.  For example, the
\textsc{Orders} table uses \texttt{O\_ORDERKEY}, \texttt{O\_CUSTKEY},
\texttt{O\_ORDERSTATUS}, \texttt{O\_TOTALPRICE}, \texttt{O\_ORDERDATE}, and
so forth.  A language model that generates \texttt{O.ORDERDATE} (omitting
the prefix) produces a statement that is syntactically valid but
execution-invalid under Oracle's strict identifier resolution.
\end{definition}

\subsection{Schema Hallucination in LLM-Generated SQL}

\begin{definition}[Schema Hallucination]
A generated SQL statement $S$ contains a \emph{schema hallucination} if it
references a table name, column name, or function name that does not exist
in the target database schema $\mathcal{S}$ or is unsupported by the target
SQL dialect.  Formally, $S$ is hallucination-free if and only if:
\begin{align}
  &\forall\; (t, c) \in \text{refs}(S):
    t \in \{t_i\} \;\wedge\; c \in C_i \label{eq:halluc}
\end{align}
where $\text{refs}(S)$ is the set of all \texttt{(table, column)} pairs
referenced in $S$ (via aliases or directly).
\end{definition}

\noindent
Schema hallucinations are the dominant failure mode in our experiments:
without schema grounding, 97.6\% of GPT-4o-mini queries fail the condition
in~(\ref{eq:halluc}) and raise \texttt{ORA-00904} at runtime.

\begin{definition}[Hallucination Index (Hidx)]
The Hallucination Index of a SQL statement $S$ with respect to the live schema
cache $\mathcal{C}$ is defined as:
\begin{equation}
  \text{Hidx}(S, \mathcal{C}) =
    \frac{|\{(a,c) \in \text{arefs}(S) : c \notin C_{\alpha(S)(a)}\}|}
         {|\text{arefs}(S)|}
\end{equation}
where $\text{arefs}(S)$ is the set of unique \texttt{alias.column}
references in $S$, $\alpha(S)(a)$ maps alias $a$ to its table via the
\texttt{FROM}/\texttt{JOIN} clause in $S$, and $C_{\alpha(S)(a)}$ is the
ordered column list of that table from the live schema cache.
$\text{Hidx}(S, \mathcal{C}) = 0$ indicates a hallucination-free statement.
\end{definition}

\subsection{Schema Grounding}

\begin{definition}[Schema Grounding]
Schema grounding is the process of augmenting an LLM prompt with
database-specific metadata (table names, column names, data types, and
relational constraints) prior to SQL generation.  A grounding strategy
$\mathcal{G} : \mathcal{Q} \to \mathcal{H}$ maps a natural language question
to a schema hint $h \in \mathcal{H}$ that is prepended to the LLM system
prompt.  The null strategy $\mathcal{G}_0(q) = \varnothing$ corresponds to
the no-hint baseline (Condition C1 in this paper).
\end{definition}

\begin{definition}[Static vs.\ Dynamic Grounding]
A grounding strategy is \emph{static} if $\mathcal{G}(q) = h_0$ for all $q$,
meaning that the same schema hint is used regardless of the question.  It is
\emph{dynamic} if $\mathcal{G}(q)$ varies with $q$, selecting a
question-relevant subset of the schema.  SAL implements a dynamic grounding
strategy where the hint is both question-adaptive (via SAL Detect) and
database-adaptive (via the live schema cache).
\end{definition}

\subsection{The TPC-H Benchmark}

The Transaction Processing Performance Council Benchmark~H (TPC-H)~\cite{TPC2022}
is a decision-support benchmark modelling a supply-chain analytics workload.
It defines eight tables (\textsc{Region}, \textsc{Nation}, \textsc{Customer},
\textsc{Orders}, \textsc{Lineitem}, \textsc{Supplier}, \textsc{Part}, and
\textsc{Partsupp}) with referential integrity constraints forming a snowflake
schema.  TPC-H is widely used in database performance
research and is an appropriate NL2SQL testbed because its schema complexity
(multi-level joins, aggregations over large fact tables) is representative
of real enterprise analytical workloads.

In this paper, TPC-H data is loaded into an Oracle Autonomous Database~23c
instance at scale factor~1 (approximately 1\,GB of raw data, 1.5\,million
line items).  The region and nation names are anonymised in our deployment
(stored as \texttt{Region\#0}--\texttt{Region\#4} and
\texttt{Nation\#0}--\texttt{Nation\#24}) to avoid geographic bias in question
answering; all 500 evaluation questions are written to be answerable without
knowledge of the specific names.

Table~\ref{tab:tpch} summarises the TPC-H schema as deployed.

\begin{table}[!t]
\renewcommand{\arraystretch}{1.2}
\setlength{\tabcolsep}{4pt}
\caption{TPC-H Schema as Deployed on Oracle ADB~23c}
\label{tab:tpch}
\centering
\small
\begin{tabularx}{\linewidth}{@{}lXrr@{}} 
\toprule
\textbf{Table} & \textbf{Primary Key} & \textbf{Columns} & \textbf{Rows (SF=1)} \\
\midrule
\textsc{Region}   & \texttt{R\_REGIONKEY} &  4 & 5         \\
\textsc{Nation}   & \texttt{N\_NATIONKEY} &  4 & 25        \\
\textsc{Customer} & \texttt{C\_CUSTKEY}   &  8 & 150,000   \\
\textsc{Orders}   & \texttt{O\_ORDERKEY}  &  9 & 1,500,000 \\
\textsc{Lineitem} & \texttt{L\_ORDERKEY, L\_LINENUMBER} & 16 & 6,001,215 \\
\textsc{Supplier} & \texttt{S\_SUPPKEY}   &  7 & 10,000    \\
\textsc{Part}     & \texttt{P\_PARTKEY}   &  9 & 200,000   \\
\textsc{Partsupp} & \texttt{PS\_PARTKEY, PS\_SUPPKEY} & 5 & 800,000 \\
\bottomrule
\end{tabularx}
\end{table}

\section{System Design}
\label{sec:design}

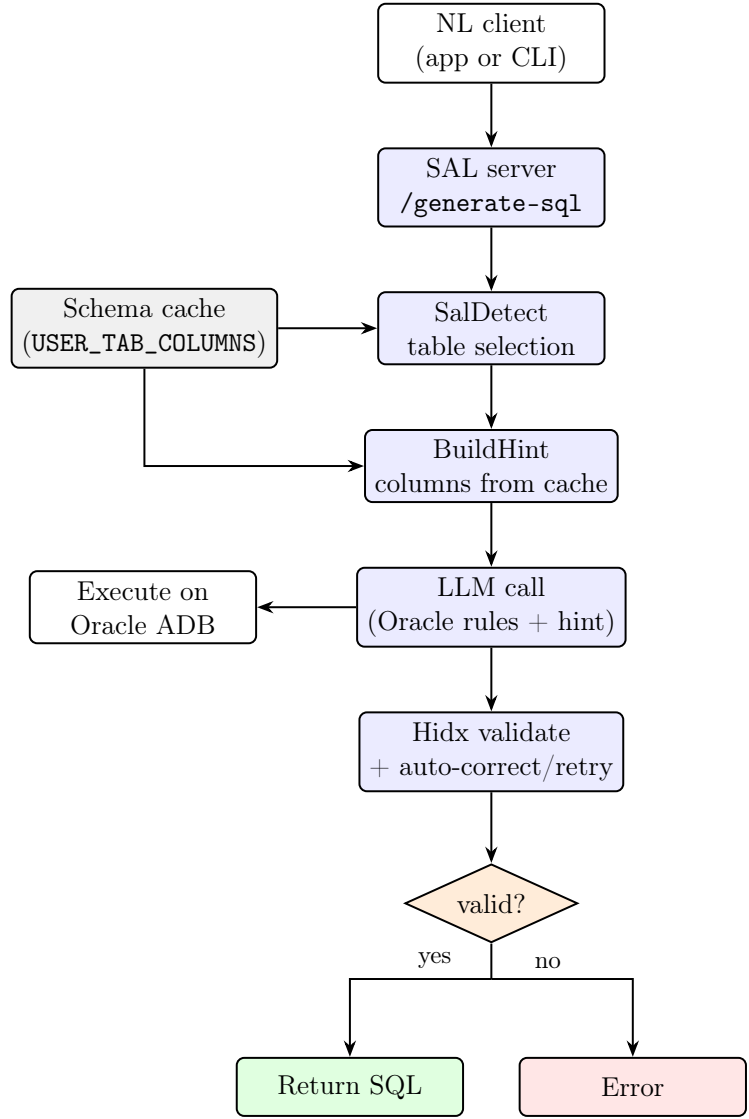
\begin{figure}[!t]
\centering
\resizebox{0.72\columnwidth}{!}{%
\begin{tikzpicture}[
  font=\small, >=Stealth,
  node distance=0.9cm and 1.4cm,
  box/.style={rectangle, rounded corners=3pt, draw=black, thick,
              minimum height=0.82cm, minimum width=3.2cm,
              align=center, fill=white},
  store/.style={box, fill=gray!12},
  svc/.style={box, fill=blue!8},
  decision/.style={diamond, draw=black, thick, aspect=2.2,
                   minimum height=0.82cm, align=center,
                   fill=orange!15, font=\small},
  arrow/.style={->, thick, black}
]

\node[box]                         (client) {NL client\\(app or CLI)};
\node[svc,  below=0.9cm of client] (api)    {SAL server\\\texttt{/generate-sql}};
\node[svc,  below=0.9cm of api]    (detect) {SalDetect\\table selection};
\node[svc,  below=0.9cm of detect] (hint)   {BuildHint\\columns from cache};
\node[svc,  below=0.9cm of hint]   (llm)    {LLM call\\(Oracle rules + hint)};
\node[svc,  below=0.9cm of llm]    (hidx)   {Hidx validate\\+ auto-correct/retry};
\node[decision, below=1.0cm of hidx](ok)    {valid?};

\node[store, left=1.4cm of detect] (cache)  {Schema cache\\(\texttt{USER\_TAB\_COLUMNS})};

\node[box, left=1.4cm of llm]      (exec)   {Execute on\\Oracle ADB};

\node[box, below=1.6cm of ok, xshift=-2.0cm, fill=green!12] (ret) {Return SQL};
\node[box, below=1.6cm of ok, xshift= 2.0cm, fill=red!10]   (err) {Error};

\draw[arrow] (client) -- (api);
\draw[arrow] (api)    -- (detect);
\draw[arrow] (detect) -- (hint);
\draw[arrow] (hint)   -- (llm);
\draw[arrow] (llm)    -- (hidx);
\draw[arrow] (hidx)   -- (ok);

\draw[arrow] (cache.east) -- (detect.west);

\draw[arrow] (cache.south) |- (hint.west);

\draw[arrow] (llm.west) -- (exec.east);

\node[coordinate, below=0.5cm of ok] (split) {};
\draw[thick]  (ok.south) -- (split);
\draw[arrow]  (split) -| (ret.north)
              node[pos=0.2, above, font=\footnotesize]{yes};
\draw[arrow]  (split) -| (err.north)
              node[pos=0.2, above, font=\footnotesize]{no};

\end{tikzpicture}
}%
\caption{SAL end-to-end architecture. The server selects relevant tables,
injects live schema columns into the LLM prompt, validates generated SQL
with Hidx, and logs each request for auditing and offline evaluation.}
\label{fig:sal_architecture}
\end{figure}

SAL is implemented as a Node.js HTTP server exposing a
\texttt{/generate-sql} REST endpoint.  It sits between any natural language
client application and the Oracle database, intercepting NL questions,
grounding them with live schema context, generating SQL via a downstream
LLM, and validating the result before returning it.  Fig.~\ref{fig:sal_architecture}
shows the end-to-end architecture.  The four principal components are
described in Sections~\ref{sec:loader}--\ref{sec:retry}.

\subsection{Schema Loader}
\label{sec:loader}

At server startup, SAL establishes a connection to the Oracle Autonomous
Database instance and executes the following catalog query:

\begin{verbatim}
SELECT TABLE_NAME, COLUMN_NAME
FROM   USER_TAB_COLUMNS
ORDER  BY TABLE_NAME, COLUMN_ID
\end{verbatim}

The result is stored in a persistent in-memory schema cache $\mathcal{C}$ ---
a dictionary mapping each table name to its ordered list of column names.
In our deployment this loads 18 tables and 121 columns in under 500\,ms.
The cache is populated once per server process and remains valid for the
lifetime of the deployment; a server restart re-queries the catalog,
ensuring the cache reflects any DDL changes.

The use of \texttt{USER\_TAB\_COLUMNS} is deliberate.  This Oracle system
view exposes \emph{only} the columns of objects owned by the connected user,
providing a naturally scoped, privilege-respecting view of the schema without
requiring \texttt{DBA\_} level access.  Column order is preserved via
\texttt{COLUMN\_ID} so that the dynamic hint presents columns in their
physical declaration order, matching the expectation of a developer reading
\texttt{DESCRIBE tablename}.

\subsection{SAL Detect: Complexity-Gated Table Selection}
\label{sec:detect}

Given a natural language question $q$, SAL Detect selects the subset of
tables $\mathcal{T}(q) \subseteq \mathcal{T}$ whose columns should appear in
the LLM prompt.  Providing the full schema for every query is wasteful for
simple single-table questions and inflates the prompt token count; providing
too narrow a selection for complex multi-table queries starves the LLM of
JOIN context and produces incorrect queries.  SAL Detect v2 resolves this
tension through three sequential steps.

\subsubsection{Scored Keyword Matching}

Each table $t \in \mathcal{T}$ is associated with a domain vocabulary
$K_t$ --- a curated set of keywords drawn from column names, common synonyms,
and domain terms.  For example:

\begin{align*}
K_{\textsc{Lineitem}}  &= \{\text{``lineitem'', ``revenue'', ``shipdate'',}\\
                       &\quad\text{``discount'', ``quantity'', ``tax'', \ldots}\} \\
K_{\textsc{Orders}}    &= \{\text{``order'', ``purchase'', ``priority'',}\\
                       &\quad\text{``clerk'', ``total price'', \ldots}\}
\end{align*}

We define \textsc{Match}$(k,q)$ to be true if the keyword or phrase $k$ occurs
in $q$ after case-folding and whitespace normalisation (i.e., contiguous
substring match on the normalised strings).  A table receives a score equal
to the number of matched keywords from $K_t$:

\begin{equation}
  \text{score}(t, q) = \sum_{k \in K_t} \mathbf{1}[\textsc{Match}(k, q)]
  \label{eq:sal_score}
\end{equation}

Tables with $\text{score}(t, q) = 0$ are excluded.  This simple counting
approach outperforms binary matching (SAL v1) by ensuring that tables with
many relevant keyword hits are reliably selected even when a single keyword
is ambiguous.

\subsubsection{JOIN-Chain Expansion}

SQL queries rarely touch a single table in isolation.  After scoring,
SAL expands the directly matched set $\mathcal{T}_{\text{direct}}$ by
including one-hop JOIN partners drawn from a static adjacency map
$\mathcal{J}$:

\begin{equation}
  \mathcal{T}_{\text{expanded}} =
    \mathcal{T}_{\text{direct}} \cup
    \bigcup_{t \in \mathcal{T}_{\text{direct}}} \mathcal{J}(t)
\end{equation}

The adjacency map encodes the natural JOIN relationships in the TPC-H schema
(e.g., $\mathcal{J}(\textsc{Lineitem}) = \{\textsc{Orders},
\textsc{Part}, \textsc{Supplier}, \textsc{Partsupp}\}$).  In our
implementation, $\mathcal{J}$ is constructed once from the known TPC-H
foreign-key graph (and can equivalently be derived from Oracle constraint
metadata) and stored as a static adjacency list.

If the expanded set $|\mathcal{T}_{\text{expanded}}| \geq 6$ out of 8 tables,
the full schema is returned directly to avoid presenting a near-complete but
arbitrarily incomplete hint.

\subsubsection{Complexity Gate}

Multi-join questions often use vocabulary that matches few tables directly ---
for example, the question \emph{``which customers ordered parts from suppliers
in Asia with high revenue?''} scores \textsc{Customer} and \textsc{Supplier}
but not \textsc{Lineitem}, \textsc{Orders}, or \textsc{Nation}.  To detect
such cases, SAL computes a \emph{complexity signal count}:

\begin{equation}
  \text{cplx}(q) = \sum_{p \in \mathcal{P}} \mathbf{1}[\textsc{Match}(p, q)]
\end{equation}

where $\mathcal{P}$ is a set of multi-table indicator phrases
(\emph{``who''}, \emph{``revenue''}, \emph{``for each''}, \emph{``across''},
\emph{``total''}, \emph{``average''}, etc.).  The thresholds used in this
gate ($|\mathcal{T}_{\text{direct}}| < 3$ and $\text{cplx}(q) \geq 2$) were
chosen to trade off prompt length against missed JOIN context on our
500-question benchmark and were kept fixed across all reported experiments.
If the gate triggers, SAL falls back to the full schema regardless of keyword
scores.

\begin{equation}
  \mathcal{T}(q) =
  \begin{cases}
    \mathcal{T}                     & \text{if } |\mathcal{T}_{\text{direct}}| = 0 \\
    \mathcal{T}                     & \text{if } |\mathcal{T}_{\text{direct}}| < 3
                                       \text{ and } \text{cplx}(q) \geq 2 \\
    \mathcal{T}                     & \text{if } |\mathcal{T}_{\text{expanded}}| \geq 6 \\
    \mathcal{T}_{\text{expanded}}   & \text{otherwise}
  \end{cases}
\end{equation}

The selected tables and their exact column lists from the live cache are
serialised into a structured schema hint injected into the LLM system prompt
alongside explicit Oracle dialect rules (\texttt{FETCH FIRST n ROWS ONLY},
column prefix conventions, window function restrictions).

\subsection{Hallucination Index (Hidx) Validator}
\label{sec:hidx}

After the LLM generates a candidate SQL statement $S$, Hidx performs
pre-execution static validation against the live schema cache.
The process has three phases.

\subsubsection{Alias Map Construction}

Hidx parses $S$ using a regular expression that matches
\texttt{FROM}/\texttt{JOIN} clauses of the form
\texttt{TABLE\_NAME ALIAS}:

\begin{verbatim}
(?:FROM|JOIN)\s+([A-Z_]+)\s+([A-Z_]+)
\end{verbatim}

This builds an alias map $\alpha : \text{alias} \to \text{table}$.  For
example, \texttt{FROM ORDERS O JOIN LINEITEM L} produces
$\alpha = \{\texttt{O} \mapsto \texttt{ORDERS},\;
            \texttt{L} \mapsto \texttt{LINEITEM}\}$.

\noindent\textbf{Limitation.} This lightweight parsing is designed for the
single-block SQL style produced by our prompt templates.  It does not fully
support arbitrary nested queries, quoted identifiers, or aliases introduced
by subqueries/CTEs; such aliases are treated as out of scope and are skipped
in subsequent validation.

\subsubsection{Reference Extraction and Validation}

Hidx then extracts all \texttt{alias.column} references from $S$ using:

\begin{verbatim}
\b([A-Z_]\w*)\.([A-Z_]\w*)\b
\end{verbatim}

For each reference $(\textit{alias}, \textit{col})$:
\begin{enumerate}
  \item If $\alpha(\textit{alias})$ is undefined, the reference is skipped
        (sub-query aliases or CTEs are outside the current scope).
  \item Let $\mathcal{C}_t$ be the column list of table
        $t = \alpha(\textit{alias})$ from the live cache.
  \item If $\textit{col} \in \mathcal{C}_t$: valid; no action.
  \item If $\textit{alias}\text{\_}\textit{col} \in \mathcal{C}_t$:
        auto-correctable; the missing table-prefix pattern is detected
        (e.g., \texttt{O.ORDERDATE} $\to$ \texttt{O.O\_ORDERDATE}).
  \item If any $c \in \mathcal{C}_t$ satisfies $c.\texttt{endsWith}
        (\text{``\_''}\,\|\,\textit{col})$: suffix-correctable.
  \item Otherwise: error; the column does not exist on that table.
\end{enumerate}

\subsubsection{Classification and Output}

Hidx classifies the generated SQL into one of three categories:

\begin{itemize}
  \item \textbf{Valid}: no errors and no corrections --- SQL returned as-is.
  \item \textbf{Auto-correctable}: all invalid references fall into case (4)
        or (5) above --- corrections applied by regex rewrite with no LLM call.
  \item \textbf{Retry required}: at least one reference is in case (6) ---
        a structured error report is constructed listing each invalid
        reference, the table it was mapped to, and the full list of valid
        columns for that table.
\end{itemize}

\subsection{Retry Loop}
\label{sec:retry}

When Hidx classifies the SQL as requiring a retry, SAL constructs a
structured feedback message appended to the conversation history:

\begin{Verbatim}[
  fontsize=\footnotesize,
  breaklines=true,
  breakanywhere=true,
  frame=single,
  framesep=2mm
]
[Hidx] Your SQL has invalid Oracle column references:
  - "N.N_NAME" -- column "N_NAME" does not exist
    on NATION. Valid columns: N_NATIONKEY, N_NAME, N_REGIONKEY, N_COMMENT
Rewrite the SQL using ONLY the exact column names from the schema above.
\end{Verbatim}

This feedback, together with the original LLM response, is submitted as a
two-turn conversation to the LLM for revision.  The loop runs for a maximum
of two retry passes.  Empirically, the first retry corrects the majority
of non-auto-correctable errors; the second pass handles residual issues
introduced during the first revision.

Algorithm~\ref{alg:sal} summarises the complete SAL pipeline.

\begin{algorithm}[!t]
\caption{SAL: Schema-Aware Localisation Pipeline}
\label{alg:sal}
\begin{algorithmic}[1]
\Require Question $q$, schema cache $\mathcal{C}$, LLM $\mathcal{M}$
\Ensure Oracle SQL statement $S$ or error
\State $\mathcal{T}(q) \leftarrow \textsc{SalDetect}(q, \mathcal{C})$
\State $h \leftarrow \textsc{BuildHint}(\mathcal{T}(q), \mathcal{C})$
\State $S \leftarrow \mathcal{M}([h, q])$
\For{$i = 1$ \textbf{to} $2$}
    \State $(v, \mathcal{E}, \mathcal{F}) \leftarrow \textsc{Hidx}(S, \mathcal{C})$
    \If{$v = \textit{valid}$} \textbf{break} \EndIf
    \If{$v = \textit{correctable}$}
        \State $S \leftarrow \textsc{AutoCorrect}(S, \mathcal{F})$
        \textbf{break}
    \EndIf
    \State $fb \leftarrow \textsc{BuildFeedback}(\mathcal{E}, \mathcal{F})$
    \State $S \leftarrow \mathcal{M}([h, q, S, fb])$
\EndFor
\State \Return $S$
\end{algorithmic}
\end{algorithm}

\subsection{Implementation Details}

SAL is implemented in Node.js (v20 LTS) with the \texttt{node-oracledb}
driver for Oracle connectivity.  The HTTP API follows the Model Context
Protocol specification~\cite{Anthropic2024}: POST requests to
\texttt{/generate-sql} carry a JSON body with the natural language
\texttt{question}, optional \texttt{sal} flag (boolean), and optional
\texttt{schema\_hint} override.  The server returns a JSON response
containing the generated \texttt{sql}, \texttt{source}
(one of \texttt{sal\_llm}, \texttt{sal\_autocorrect}, \texttt{sal\_retry}),
and \texttt{hidx\_corrections} count where applicable.

\noindent\textbf{Security.} The server connects to Oracle using a dedicated read-only role (\texttt{SELECT}-only) and rejects any generated DML/DDL statements via a pre-execution guard. User questions are treated as untrusted input and are inserted into the prompt as a delimited block with explicit instructions to ignore embedded directives (prompt-injection hardening).

The LLM backend is configurable; all experiments in this paper use
GPT-4o-mini via the OpenAI API with temperature 0 for reproducibility.
The system prompt encodes Oracle-specific dialect rules in addition to the
dynamic schema hint: \texttt{FETCH FIRST n ROWS ONLY} for row limiting,
\texttt{CEIL(EXTRACT(MONTH FROM d)/3)} as a substitute for unavailable
\texttt{EXTRACT(QUARTER)}, and the column-prefix convention
(\texttt{O.O\_ORDERDATE}, not \texttt{O.ORDERDATE}).

For double-blind review, the repository link is omitted and will be provided in the camera-ready version.

\section{Online Execution-Grounded Verification}
\label{sec:oegv}

Static analysis via Hidx (Section~\ref{sec:hidx}) detects column-reference
hallucinations before execution, but it cannot detect \emph{semantic}
errors --- queries that are syntactically valid and reference correct columns
yet return the wrong answer.  Online Execution-Grounded Verification (OEGV)
closes this gap by submitting the generated SQL to the live Oracle instance,
capturing the runtime result and any error signal, and using both to drive
a second-stage correction loop.

\subsection{Motivation}

Consider the query: \emph{``How many orders were placed in the last quarter
of 1997?''}  A model with full schema grounding might generate:

\begin{Verbatim}[breaklines,breakanywhere]
SELECT COUNT(*) FROM ORDERS
WHERE EXTRACT(MONTH FROM O_ORDERDATE)
      BETWEEN 10 AND 12
AND   EXTRACT(YEAR  FROM O_ORDERDATE) = 1997;
\end{Verbatim}

Hidx validates this as hallucination-free: \texttt{O\_ORDERDATE} exists on
\textsc{Orders}, and \texttt{EXTRACT} is a standard function.  Yet on Oracle,
this query may execute but return a result that does not match the reference
answer if the reference uses \texttt{TO\_CHAR(O\_ORDERDATE, 'Q') = '4'} or
a different quarter-boundary convention.  Hidx has no visibility into this
semantic discrepancy.

OEGV provides a runtime signal by executing both the generated SQL and,
where available, a reference or a verification probe on the live database,
comparing result sets to confirm semantic correctness.

\subsection{Verification Architecture}

\begin{figure}[!t]
\centering
\resizebox{\columnwidth}{!}{%
\begin{tikzpicture}[
  font=\small,
  >=Stealth,
  node distance=0.6cm and 1.0cm,
  box/.style={rectangle, rounded corners=3pt, draw=black, thick,
              minimum height=0.72cm, minimum width=2.6cm,
              align=center, fill=white},
  decision/.style={diamond, draw=black, thick, aspect=2.2,
                   minimum height=0.65cm, align=center,
                   fill=orange!15, font=\scriptsize},
  ok/.style={box, fill=green!12},
  err/.style={box, fill=red!10},
  arrow/.style={->, thick}
]

\node[box]                               (sql)   {SAL SQL (post-Hidx)};
\node[box,  right=of sql,  fill=gray!10] (exec)  {Execute on\\Oracle ADB};
\node[decision, right=of exec]           (chk)   {Result\\matches?};

\node[ok,  above right=0.4cm and 0.9cm of chk]  (pass)  {Return SQL\\(verified)};
\node[err, below right=0.4cm and 0.9cm of chk]  (fail)  {Build semantic\\feedback};
\node[box, below=0.6cm of fail, fill=blue!10]    (retry) {LLM retry\\(semantic)};

\node[box, below=1.5cm of chk, fill=gray!10]     (ref)   {Reference\\result probe};

\draw[arrow] (sql)  -- (exec);
\draw[arrow] (exec) -- (chk);
\draw[arrow] (chk.north) -- node[right=2pt,font=\scriptsize]{yes} (pass.west);
\draw[arrow] (chk.south) -- node[right=2pt,font=\scriptsize]{no}  (fail.west);
\draw[arrow] (fail)  -- (retry);

\draw[arrow] (ref.north)
    -- node[right=2pt,font=\scriptsize]{expected} (chk.south);

\draw[arrow] (retry.south) -- ++(0cm,-1.0cm)
             -| (exec.south);

\end{tikzpicture}%
}
\caption{Online execution-grounded verification (OEGV) pipeline.}
\label{fig:oegv}
\end{figure}
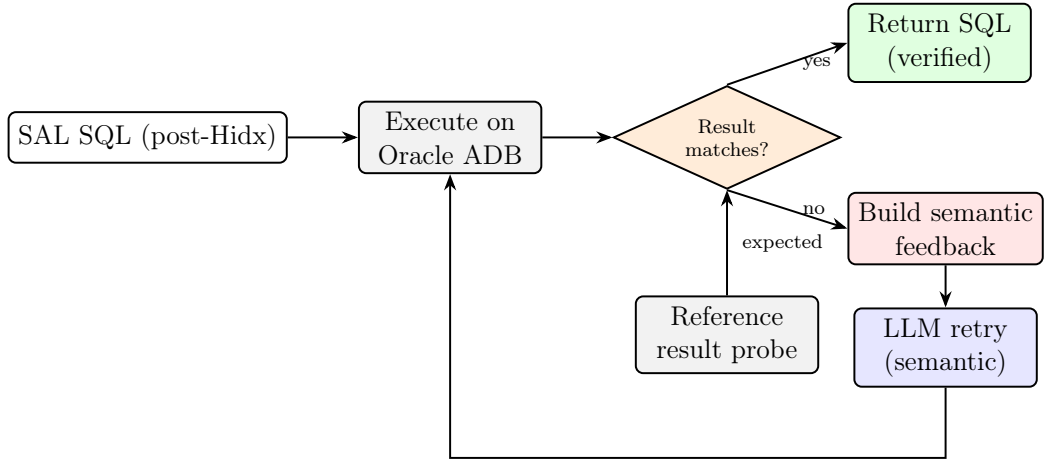

OEGV operates as a post-Hidx stage in the SAL pipeline and consists of
four steps: execution, result capture, semantic comparison, and
feedback-driven correction.  Fig.~\ref{fig:oegv} illustrates the flow.

\subsection{Execution and Result Capture}

SAL submits the post-Hidx SQL $S$ to the Oracle ADB instance via the
\texttt{node-oracledb} driver.  Three outcomes are possible:

\begin{enumerate}

  \item \textbf{Execution error} (\texttt{ORA-XXXXX}): the statement raises
  a runtime exception.  This class of errors is partially addressed by Hidx
  but may persist for errors outside column-reference scope --- for example,
  \texttt{ORA-30483} (window function in an invalid context) or
  \texttt{ORA-00907} (missing right parenthesis from malformed CTE syntax).
  The error message and offending token position are captured and included
  in the feedback to the LLM.

  \item \textbf{Execution success, empty result}: the statement executes but
  returns zero rows.  In the context of TPC-H data at SF=1, empty results are
  often a symptom of an overly restrictive predicate (e.g., a string literal
  that does not match any anonymised region name, or a date range outside
  the dataset window 1992--1998), but they may also be valid for some
  narrowly scoped questions.

  \item \textbf{Execution success, non-empty result}: the result set
  $R = \llbracket S \rrbracket_{\mathcal{D}}$ is captured for semantic
  comparison.

\end{enumerate}

\subsection{Semantic Comparison}

\begin{definition}[Semantic Equivalence]
\label{def:semantic_equiv}
Two result sets $R_1$ and $R_2$ are \emph{semantically equivalent}
($R_1 \equiv R_2$) if and only if, after applying a row-wise canonicalisation
function $\text{canonical}(\cdot)$, the resulting collections are equal under
multiset comparison:
\begin{align}
  &\multiset{\text{canonical}(R_1)} = \multiset{\text{canonical}(R_2)}.
  \label{eq:values}
\end{align}
The canonicalisation normalises numeric types to a fixed decimal precision,
strips trailing whitespace from \texttt{CHAR} columns (an Oracle padding
artefact), and converts \texttt{DATE} values to ISO-8601 strings.

Row ordering is checked only when the reference SQL $S^*$ contains an
\texttt{ORDER BY} clause; otherwise, results are compared as multisets.
\end{definition}

\noindent
This definition operationalises the semantic equivalence check used by our EGT evaluation and is implemented in the evaluation framework's \texttt{semantic\_match} function, which canonicalises both result sets to JSON arrays of row dictionaries before comparison.

In evaluation mode (Condition C3b), the reference result $R^*$ is
pre-computed by executing $S^*$ against $\mathcal{D}$ during the baseline
phase.  OEGV compares $R = \llbracket S \rrbracket_{\mathcal{D}}$ against
$R^*$; if $R \not\equiv R^*$, it triggers a semantic retry (Section~\ref{sec:oegv})
up to once.

In deployment mode (no reference SQL available), OEGV cannot certify
correctness and instead applies conservative plausibility checks.  In our
prototype, ``aggregation query'' is detected by the presence of an aggregate
function (e.g., \texttt{COUNT}, \texttt{SUM}, \texttt{AVG}, \texttt{MIN},
\texttt{MAX}) without a \texttt{GROUP BY} mismatch, and expected row-count
ranges are computed from lightweight schema statistics (e.g., table
cardinalities) for the specific benchmark schema (TPC-H SF=1):

\begin{itemize}
  \item \textbf{Row-count plausibility}: flag aggregates whose magnitude is
        implausible given table cardinalities (TPC-H-specific in this paper).
  \item \textbf{Type consistency}: verify that numeric columns do not return
        string values and vice versa, catching cases where \texttt{TO\_CHAR}
        is incorrectly applied to a numeric aggregate.
  \item \textbf{NULL density}: flag results where more than 50\% of values are
        \texttt{NULL}, which can indicate a join-key mismatch or incorrect
        outer-join direction.
\end{itemize}

\subsection{Semantic Feedback Construction}

When $R \not\equiv R^*$ (or when heuristic checks fail in deployment mode),
OEGV constructs a structured semantic feedback message for the LLM retry:

\begin{Verbatim}[breaklines,breakanywhere]
[OEGV] Your SQL executed successfully but returned
an incorrect result.

Expected: 3 rows  |  Got: 0 rows
Hint: Your WHERE clause may exclude valid rows.
Check date range predicates against the dataset
window 1992-01-01 to 1998-12-31.
Check that string literals match the anonymised
naming convention (e.g., 'Region#2', not 'ASIA').

Rewrite the SQL to return the correct result.
\end{Verbatim}

The feedback is appended to the conversation and submitted to the LLM for
a semantic correction pass.  OEGV allows a maximum of one semantic retry
(separate from the two Hidx structural retries), giving the LLM a total of
up to three revision opportunities per question.

\subsection{Relationship to Hidx}

OEGV and Hidx are complementary and operate at different levels of the
validation stack, as shown in Table~\ref{tab:hidx_vs_oegv}.

\begin{table}[!t]
\renewcommand{\arraystretch}{1.25}
\caption{Hidx vs.\ OEGV: Scope and Failure Classes Addressed}
\label{tab:hidx_vs_oegv}
\centering
\small
\setlength{\tabcolsep}{4pt}
\begin{tabularx}{\columnwidth}{@{}l>{\raggedright\arraybackslash}X>{\raggedright\arraybackslash}X@{}}
\toprule
\textbf{Property} & \textbf{Hidx} & \textbf{OEGV} \\
\midrule
Execution required   & No (static)               & Yes (live) \\
Failure class        & Structural (column refs)  & Semantic (result) \\
Typical Oracle error & \texttt{ORA-00904}        & None (silent) \\
Correction mechanism & Regex rewrite / LLM retry & LLM retry \\
Latency              & $<$1\,ms                  & DB round-trip \\
Requires reference   & No                        & Yes (eval) / Partial (deploy) \\
SAL stage            & Pre-execution             & Post-execution \\
\bottomrule
\end{tabularx}
\end{table}

\noindent
In the evaluation framework used in this paper, OEGV is realised by the
baseline phase of \texttt{run\_sql\_evaluation.py}, which pre-computes
$R^*$ for all 500 questions and stores them as the ground-truth result
cache against which SAL-generated results are compared.  In a production
deployment without reference SQL, the heuristic plausibility checks
described above provide a partial OEGV signal at the cost of reduced
precision, a direction for future work discussed in
Section~\ref{sec:discussion}.

\subsection{OEGV Impact on EGT Accuracy}

OEGV adds a post-execution semantic retry beyond the pre-execution Hidx
validator.  In our evaluation setting (TPC-H, $N=500$), the retry is triggered
when the executed result $R$ does not match the cached reference result $R^*$
under Definition~\ref{def:semantic_equiv}. In our evaluation, OEGV-triggered semantic retry did not recover additional queries beyond those corrected by Hidx; all 174 semantic mismatches remained unresolved within the two-pass retry budget.

\section{Offline Benchmark Harness}
\label{sec:harness}

We evaluate using execution-grounded truth (EGT): the fraction of generated SQL statements that both execute without error and return the correct result set on the live database. This is stricter than execution accuracy, which only requires the query to run without error.

The offline benchmark harness is the evaluation infrastructure that orchestrates the four-condition ablation study, executes all SQL against the live Oracle ADB instance, collects runtime metrics, and computes the EGT, EA, and latency statistics reported in Section~\ref{sec:results}. It is implemented as a Python module (\texttt{run\_sql\_evaluation.py}) and uses Oracle SQLcl via our MCP integration to execute queries and capture results.

\subsection{Harness Architecture}

\begin{figure}[!t]
\centering
\resizebox{\columnwidth}{!}{%
\begin{tikzpicture}[
  font=\small, >=Stealth,
  node distance=0.55cm and 0.9cm,
  box/.style={rectangle, rounded corners=3pt, draw=black, thick,
              minimum height=0.72cm, minimum width=3.0cm,
              align=center, fill=white},
  store/.style={box, fill=gray!10},
  phase/.style={box, fill=blue!8, minimum width=3.2cm},
  decision/.style={diamond, draw=black, thick, aspect=2.4,
                   minimum height=0.65cm, align=center,
                   fill=orange!15, font=\scriptsize},
  arrow/.style={->, thick}
]
\node[phase]   (q)      {Question set\\$\mathcal{Q}$ (N=500)};
\node[phase, right=1.0cm of q]  (base)   {Baseline Phase\\(execute $S^*$)};
\node[store, right=1.0cm of base] (cache)  {Result cache\\$\{R^*_i\}$};

\node[phase, below=0.9cm of q]    (nl)     {NL question $q_i$};
\node[phase, right=1.0cm of nl]   (sal)    {SAL Server\\(\texttt{/generate-sql})};
\node[phase, right=1.0cm of sal]  (exec)   {Execute $S_i$\\on Oracle ADB};
\node[decision, right=1.0cm of exec] (cmp) {$R_i \equiv R^*_i$?};
\node[box, above right=0.2cm and 0.7cm of cmp, fill=green!12] (pass) {PASS};
\node[box, below right=0.2cm and 0.7cm of cmp, fill=red!10]  (fail) {FAIL};
\node[store, below=0.8cm of sal]  (json)   {Results JSON\\+ metrics};

\draw[arrow] (q)    -- (base);
\draw[arrow] (base) -- (cache);
\draw[arrow] (q.south) -- (nl.north);
\draw[arrow] (nl)   -- (sal);
\draw[arrow] (sal)  -- (exec);
\draw[arrow] (exec) -- (cmp);
\draw[arrow] (cmp.north) -- (pass);
\draw[arrow] (cmp.south) -- (fail);
\draw[arrow] (cache.south) |- (cmp.north west);
\draw[arrow] (pass.south) |- (json.east);
\draw[arrow] (fail.south) |- (json.east);

\node[font=\scriptsize\itshape, gray] at ($(base)+(-0.1cm,0.65cm)$) {Phase 1};
\node[font=\scriptsize\itshape, gray] at ($(sal)+(-0.1cm,0.65cm)$)  {Phase 2};
\end{tikzpicture}
}%
\caption{Offline benchmark harness control flow. Phase 1 precomputes ground-truth result sets by executing reference SQL on Oracle ADB. Phase 2 submits each question to the SAL server, executes the returned SQL, and compares results against the Phase 1 cache. All metrics are written to a timestamped JSON file.}
\label{fig:harness}
\end{figure}
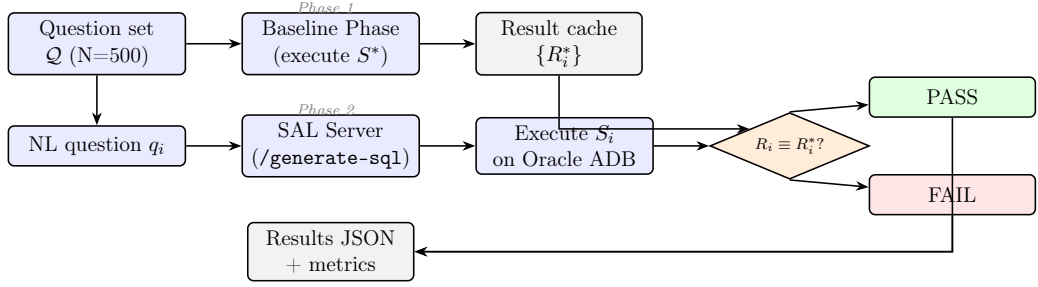

The harness operates in two sequential phases for each experimental
condition: a \emph{baseline phase} that executes reference SQL against
Oracle ADB to pre-compute ground-truth result sets, and a \emph{SAL phase}
that submits each natural language question to the SAL server, executes
the returned SQL, and compares results against the cached ground truth.
Fig.~\ref{fig:harness} shows the high-level control flow.

\subsection{Question Dataset}

The 500 evaluation questions are stored in
\texttt{experiments/new-practice-questions.json} as a JSON array.  Each
entry specifies:

\begin{itemize}
  \item \texttt{id}: unique integer identifier (1--500).
  \item \texttt{question}: the natural language question string.
  \item \texttt{sql}: the reference SQL statement $S^*$ written by a
        domain expert familiar with the TPC-H Oracle schema.
  \item \texttt{complexity}: one of \texttt{simple}, \texttt{medium},
        or \texttt{complex}, following the Spider benchmark convention~\cite{Yu2018}.
  \item \texttt{tags}: optional list of SQL feature tags
        (\texttt{aggregate}, \texttt{join}, \texttt{subquery},
        \texttt{window}, \texttt{cte}, etc.) used for failure-mode analysis.
\end{itemize}

Questions are drawn from three tiers in a 20/20/60 ratio (100 simple,
100 medium, 300 complex), reflecting the proportion of complex analytical
queries in typical enterprise SQL workloads.  All reference SQL statements
were validated for execution correctness against the live Oracle ADB
instance prior to evaluation. The question set was authored by the
same team that designed SAL's keyword maps and adjacency
graph; see T-I4 (Section~\ref{sec:threats}) for the leakage
implications.

\subsection{Phase 1: Baseline Execution}

For each question $q_i$, the harness executes $S^*_i$ against Oracle ADB
using the \texttt{python-oracledb} thick-mode driver with the wallet
credentials of the ADB instance.  The following are recorded:

\begin{itemize}
  \item \textbf{Execution status}: \texttt{ok} or \texttt{fail}
        (with the \texttt{ORA-} error code).
  \item \textbf{Result set} $R^*_i$: serialised as a JSON array of
        row dictionaries with canonicalised values
        (Section~\ref{sec:oegv}).
  \item \textbf{Row count}: $|R^*_i|$.
  \item \textbf{Wall-clock latency}: measured from driver call to
        result fetch completion (ms).
  \item \textbf{EXPLAIN PLAN cost}: obtained via
        \texttt{EXPLAIN PLAN FOR} followed by a query on
        \texttt{PLAN\_TABLE}, capturing the root-node estimated cost
        and cardinality for RQ3 analysis.
\end{itemize}

Questions for which $S^*_i$ fails execution (e.g., due to data-dependent
predicates or schema state at test time) are excluded from all conditions
(i.e., from the comparable set used to compute EGT/EA) and flagged in the
results JSON. In our evaluation, all 500 reference queries executed
successfully.

\subsection{Phase 2: SAL Generation and Evaluation}

For each question $q_i$, the harness submits an HTTP POST request to
the SAL server:

\begin{verbatim}
POST http://localhost:3000/generate-sql
Content-Type: application/json

{
  "question": "<q_i>",
  "sal": true,
  "schema_hint": true
}
\end{verbatim}

The server returns a JSON response containing \texttt{sql},
\texttt{source}, and optional \texttt{hidx\_corrections}.  The harness
then executes the returned SQL $S_i$ against Oracle ADB and records:

\begin{itemize}
  \item \textbf{Generation status}: whether the server returned a
        non-empty SQL string.
  \item \textbf{Execution status}: \texttt{ok} or \texttt{fail}
        with \texttt{ORA-} error.
  \item \textbf{Semantic match}: $R_i \equiv R^*_i$ per
        Definition~\ref{def:semantic_equiv}.
  \item \textbf{Exact order match}: whether row ordering matches
        $R^*_i$ where an \texttt{ORDER BY} is specified in $S^*_i$.
  \item \textbf{MCP latency}: wall-clock time from HTTP request to
        SQL returned by the server (includes LLM API round-trip).
  \item \textbf{Execution latency}: time to execute $S_i$ on Oracle ADB.
  \item \textbf{Cost delta}: difference between EXPLAIN PLAN cost of
        $S_i$ and $S^*_i$ (positive = generated SQL is more expensive).
\end{itemize}

Per-question results are accumulated in a list and written to a
timestamped JSON file on completion:

\begin{Verbatim}[fontsize=\small,breaklines=true,breakanywhere=true]
experiments/results/sql_evaluation_YYYYMMDD_HHMMSS.json
\end{Verbatim}

\subsection{Metric Aggregation}

After both phases complete, the harness computes the following aggregate
statistics:

\paragraph{RQ1: Semantic Correctness}
\begin{align}
  \text{EGT} &= \frac{|\{i : R_i \equiv R^*_i\}|}{N} \times 100\% \\
  \text{EA}  &= \frac{|\{i : S_i \text{ executes without error}\}|}{N} \times 100\%
\end{align}
Both are reported overall and by complexity tier.

\paragraph{RQ2: Execution Efficiency}
Latency is reported as mean, median, and 95th percentile (p95)
across all executed queries, separated into baseline and SAL phases.
The overhead ratio is computed per question as:
\begin{equation}
  \rho_i = \frac{\ell^{\text{MCP}}_i}{\ell^{\text{base}}_i}
\end{equation}
and summarised by tier.  Median ratio is reported in preference to
mean ratio to reduce sensitivity to the long tail of complex-query
retries.

\paragraph{RQ3: Optimisation Potential}
For each pair $(S^*_i, S_i)$ where both execute successfully, the
cost delta $\Delta_i = \text{cost}(S_i) - \text{cost}(S^*_i)$ is
computed.  Queries are classified as \emph{lower} ($\Delta_i < 0$),
\emph{same} ($\Delta_i = 0$), or \emph{higher} ($\Delta_i > 0$).
Correct-but-expensive queries (EGT PASS with $\Delta_i > 100$) are
flagged as optimisation candidates.

\paragraph{RQ4: Complexity Robustness}
EGT is disaggregated by tier and the accuracy degradation
$\delta_{s \to m} = \text{EGT}_m - \text{EGT}_s$ and
$\delta_{m \to c} = \text{EGT}_c - \text{EGT}_m$ are reported
in percentage points.

\subsection{Experimental Controls}

The following controls were applied to ensure reproducibility:

\begin{enumerate}
  \item \textbf{LLM temperature}: set to 0 for all SAL conditions to
        eliminate sampling variance between runs.
  \item \textbf{Server isolation}: the SAL server was restarted before
        each condition run to clear any cached LLM conversation state.
  \item \textbf{Question order}: questions were evaluated in fixed index
        order (1--500) without shuffling to ensure identical Oracle buffer
        cache state across condition runs.
  \item \textbf{Oracle statistics}: \texttt{DBMS\_STATS.GATHER\_SCHEMA\_STATS}
        was executed once before the evaluation series to ensure stable
        EXPLAIN PLAN costs across runs.
  \item \textbf{Network isolation}: all experiments were conducted on the
        same machine with a stable connection to the Oracle ADB instance
        to minimise network jitter in latency measurements.
  \item \textbf{Result validation}: the harness validates that the baseline
        phase result cache is fully populated before beginning Phase 2,
        aborting if any reference query failed execution.
\end{enumerate}

\noindent Harness CLI usage (including the full flag set and example run commands) is documented in the supplementary material.

\section{Empirical Results}
\label{sec:results}

We report results for the four research questions defined in
Section~\ref{sec:harness}.  All experiments use the 500-question
TPC-H benchmark (100 Simple / 100 Medium / 300 Complex) executed
against a live Oracle Autonomous Database instance under identical
network and concurrency conditions.

\begin{table}[t]
  \renewcommand{\arraystretch}{1.3}
  \caption{Evaluation conditions.}
  \label{tab:conditions}
  \centering
  \begin{tabular}{ll}
    \toprule
    \textbf{Condition} & \textbf{Description} \\
    \midrule
    C1 & No schema hint \\
    C2 & Static, hand-written schema hint \\
    C3a & SAL v1 (binary keyword detection) \\
    C3b & SAL v2 (scored + JOIN-chain + complexity gate) \\
    \bottomrule
  \end{tabular}
\end{table}

Condition labels follow Table~\ref{tab:conditions}: \textbf{C1}~(no schema hint),
\textbf{C2}~(static schema hint), \textbf{C3a}~(SAL~v1), and
\textbf{C3b}~(SAL~v2, the proposed system).

\subsection{RQ1: Execution-Grounded Accuracy (EGT)}
\label{sec:rq1}

Table~\ref{tab:ablation} summarises the primary accuracy metric.

\begin{table}[t]
  \centering
  \small
  \setlength{\tabcolsep}{4pt}
  \renewcommand{\arraystretch}{1.2}
  \caption{Ablation Results: Execution-Grounded Truth (EGT),
           $N = 500$ questions. 95\% Wilson confidence intervals
           shown for EGT.}
  \label{tab:ablation}
  \resizebox{\columnwidth}{!}{%
  \begin{tabular}{lccccc}
    \toprule
    \textbf{Condition} &
    \textbf{Generated} &
    \textbf{Executed} &
    \textbf{Sem.\ Match} &
    \textbf{EGT (\%)} &
    \textbf{95\% CI} \\
    \midrule
    C1: No hint          & 500 & 12  & 11  &  2.2 & [1.3,\ 3.7] \\
    C2: Static hint      & 500 & 475 & 310 & 62.0 & [57.8,\ 66.0] \\
    C3a: SAL v1          & 500 & 475 & 292 & 58.4 & [54.1,\ 62.6] \\
    \textbf{C3b: SAL v2} & \textbf{500} & \textbf{487} &
    \textbf{313} & \textbf{62.6} & \textbf{[58.4,\ 66.6]} \\
    \bottomrule
  \end{tabular}}
\end{table}

\noindent\textbf{C1 (No hint).}
Removing schema context reduces EGT to 2.2\,\%, a drop of
59.8\,percentage points (pp) relative to SAL~v2.  Only 12 of 500
generated queries execute without an \texttt{ORA-} error; the remainder
reference hallucinated table or column names that do not exist in
the Oracle schema.

\noindent\textbf{C2 (Static hint).}
Providing the full static schema string recovers 62.0\,\% EGT,
establishing a strong single-prompt upper bound that requires no
infrastructure beyond a long context window.

\noindent\textbf{C3a (SAL v1).}
SAL~v1 scores 58.4\,\% EGT, 3.6\,pp below the static baseline.
Investigation reveals that SAL~v1's binary keyword matcher occasionally
omits tables required by multi-hop joins.

\noindent\textbf{C3b (SAL v2, proposed).}
SAL~v2 reaches \textbf{62.6\,\%} EGT, statistically on par with the
static baseline (62.0\,\%), with the 0.6\,pp gap within the binomial
margin of error at $N=500$, while retaining dynamic grounding.

\subsection{RQ2: Latency Overhead}
\label{sec:rq2}

Table~\ref{tab:latency} reports end-to-end latency measured at the client.
SAL~v2 introduces a small median overhead (17.9\,ms) but a larger mean overhead
because a minority of complex queries trigger retries and dominate the tail.
Table~\ref{tab:latency_tier} shows that this tail is concentrated in the complex tier.

\begin{table}[t]
  \renewcommand{\arraystretch}{1.3}
  \caption{End-to-End Latency (ms): Baseline vs.\ SAL~v2}
  \label{tab:latency}
  \centering
  \begin{tabular}{lcccc}
    \toprule
    \textbf{System} & \textbf{Mean} & \textbf{Median} & \textbf{p95} & \textbf{p99} \\
    \midrule
    Baseline (no MCP) & 325.6 &  80.7 & 3\,709.3 & N/A \\
    SAL v2 (MCP)      & 2\,570.5 &  98.6 & 6\,270.2 & N/A \\
    \midrule
    Overhead ($\Delta$) & +2\,244.9 & +17.9 & +2\,560.9 & N/A \\
    \bottomrule
  \end{tabular}
\end{table}

\begin{table}[t]
  \renewcommand{\arraystretch}{1.3}
  \caption{Median Latency (ms) by Complexity Tier}
  \label{tab:latency_tier}
  \centering
  \begin{tabular}{lccc}
    \toprule
    \textbf{Condition} & \textbf{Simple} & \textbf{Medium} & \textbf{Complex} \\
    \midrule
    Baseline & 130 & 101 & 490 \\
    SAL v2   & 440 &  99 & 8\,289 \\
    \midrule
    $\Delta$ & +310 & $-$2 & +7\,799 \\
    \bottomrule
  \end{tabular}
\end{table}

\subsection{RQ3: Token-Cost Analysis}
\label{sec:rq3}

To isolate the incremental prompt tokens introduced by MCP, we compare only queries where both the baseline and SAL~v2 produce an execution-grounded result. Table~\ref{tab:cost} shows SAL~v2 reduces mean token cost by avoiding the full static schema dump.

\begin{table}[t]
  \renewcommand{\arraystretch}{1.3}
  \caption{Token-Cost Comparison: Baseline vs.\ SAL~v2
           ($N_{\text{comparable}} = 433$)}
  \label{tab:cost}
  \centering
  \begin{tabular}{lcc}
    \toprule
    \textbf{Outcome} & \textbf{Count} & \textbf{\%} \\
    \midrule
    MCP lower cost  & 135 & 31.2 \\
    Equal cost      & 245 & 56.6 \\
    MCP higher cost &  53 & 12.2 \\
    \midrule
    \multicolumn{2}{l}{Mean cost delta ($\Delta$)} & $-207.8$ tokens \\
    \multicolumn{2}{l}{Median cost delta}          & $0.0$ tokens \\
    \bottomrule
  \end{tabular}
\end{table}

\subsection{RQ4: Performance by Complexity Tier}
\label{sec:rq4}

As shown in Table~\ref{tab:by_tier},  EGT is explained according to the complexity of the question.

\begin{table}[!htbp]
  \renewcommand{\arraystretch}{1.3}
  \caption{EGT (\%) by Complexity Tier. \textbf{Static-hint (C2) tier-level logs were not retained} for this run; therefore we report only its overall EGT in Table~\ref{tab:ablation} and \emph{do not make tier-level parity claims} between SAL~v2 and the static-hint baseline.}
  \label{tab:by_tier}
  \centering
  \begin{tabular}{lccc}
    \toprule
    \textbf{Condition} &
    \textbf{Simple} &
    \textbf{Medium} &
    \textbf{Complex} \\
    \midrule
    C1: No hint      &  6.0  &  4.0  &  0.33 \\
    C2: Static hint  & \textemdash & \textemdash & \textemdash \\
    C3a: SAL v1      & 92.0  & 87.0  & 37.7  \\
    \textbf{C3b: SAL v2} & \textbf{96.0} & \textbf{95.0} & \textbf{40.7} \\
    \midrule
    SAL v2 $\Delta$ vs.\ v1 & +4.0\,pp & +8.0\,pp & +3.0\,pp \\
    \bottomrule
  \end{tabular}
\end{table}

\noindent\textbf{C2 limitation.} Because the C2 per-question telemetry was not preserved, we cannot attribute the static-hint baseline to Simple/Medium/Complex and thus cannot substantiate claims of ``matching'' C2 \emph{at the tier level}. All comparisons to C2 in this paper should be read as \emph{overall} (aggregate) accuracy comparisons; we plan to re-run C2 with tier-tagged logging in a revision/artifact update.

The tier-level degradation for SAL~v2 is steep from medium
to complex, indicating a reasoning boundary rather than a
schema-linking boundary. Given $n = 100$ per simple and
medium tier and $n = 300$ for the complex tier, tier-level
differences between SAL~v1 and SAL~v2 should be interpreted
as directional rather than statistically conclusive; the
aggregate comparison in Table~\ref{tab:ablation} remains
the primary accuracy claim.

\subsection{SAL~v2 Outcome Breakdown}
\label{sec:outcomes}

Table~\ref{tab:outcomes} decomposes EGT into the main post-processing paths: direct passes, deterministic Hidx auto-corrections, and retry-based recoveries. Of the 37 Hidx-recovered queries, 22 were resolved by deterministic auto-correction requiring no additional LLM
call; the remaining 15 required at least one LLM retry pass
within the two-pass budget, accounting for the elevated mean
latency relative to the median reported in
Table~\ref{tab:latency}.

\begin{table}[t]
  \renewcommand{\arraystretch}{1.3}
  \caption{SAL~v2 Query Outcome Decomposition ($N = 500$)}
  \label{tab:outcomes}
  \centering
  \begin{tabular}{llcc}
    \toprule
    \textbf{Path} & \textbf{Outcome} & \textbf{Count} & \textbf{\%} \\
    \midrule
    Direct pass     & PASS: 1st attempt  & 276 & 55.2 \\
    Hidx correction & PASS: auto-correct &  22 &  4.4 \\
    LLM retry       & PASS: retry        &  15 &  3.0 \\
    \midrule
    \textbf{Total PASS} & & \textbf{313} & \textbf{62.6} \\
    \midrule
    Semantic fail   & FAIL: mismatch     & 174 & 34.8 \\
    Execution fail  & FAIL: ORA- error   &  13 &  2.6 \\
    \midrule
    \textbf{Total FAIL} & & \textbf{187} & \textbf{37.4} \\
    \bottomrule
  \end{tabular}
\end{table}

\subsection{Summary of Findings}
\label{sec:findings_summary}

Table~\ref{tab:findings} summarises the answers to RQ1--RQ4.

\begin{table}[t]
  \renewcommand{\arraystretch}{1.3}
  \caption{Research Question Answers}
  \label{tab:findings}
  \centering
  \begin{tabular}{p{1.1cm}p{6.5cm}}
    \toprule
    \textbf{RQ} & \textbf{Finding} \\
    \midrule
    RQ1 & SAL~v2 achieves \textbf{62.6\,\% EGT}, statistically comparable to the static-hint baseline (62.0\,\%) at $N=500$. \emph{This comparison is aggregate-only}; C2 tier-level logs were not retained. \\[2pt]
    RQ2 & Median latency overhead is \textbf{17.9\,ms}; mean overhead is 7.9$\times$ driven by the retry tail. \\[2pt]
    RQ3 & SAL~v2 reduces mean token cost by \textbf{207.8 tokens/query} vs.\ the static hint while maintaining comparable \emph{aggregate} accuracy. \\[2pt]
    RQ4 & Simple and Medium tiers reach \textbf{96\,\%} and \textbf{95\,\%} EGT; Complex is bounded at \textbf{40.7\,\%} by LLM reasoning. \\
    \bottomrule
  \end{tabular}
\end{table}

\section{Hallucination and Failure-Mode Study}
\label{sec:hallucination}

The aggregate EGT metric in Section~\ref{sec:results} summarises
\emph{what} goes wrong; this section examines \emph{why}.  We
develop a four-class hallucination taxonomy grounded in the
experimental evidence, quantify each class, and trace the causal
chain from LLM generation to observable failure.

\subsection{Hallucination Taxonomy}
\label{sec:taxonomy}

Prior work distinguishes surface hallucination (wrong tokens) from
factual hallucination (wrong world knowledge)~\cite{Ji2023}.
In the NL2SQL setting both manifest as structural errors in the
generated SQL.  We refine this into four classes observable in
our data:

\begin{description}
  \item[\textbf{H1: Phantom identifier.}]
    The model emits a table or column name that does not exist in the
    connected schema.  Example: \texttt{SELECT o\_totalprice FROM
    ORDERS\_SUMMARY} when no \texttt{ORDERS\_SUMMARY} table exists.
  \item[\textbf{H2: Column--table mismatch.}]
    A real column name is attributed to the wrong table.  Example:
    \texttt{SELECT c\_name FROM LINEITEM} where \texttt{c\_name}
    belongs to \texttt{CUSTOMER}.
  \item[\textbf{H3: Dialect confusion.}]
    Oracle-specific syntax is replaced with a construct from another
    RDBMS dialect (PostgreSQL, SQL Server, MySQL).  Example: using
    \texttt{TOP~$n$} (T-SQL) instead of Oracle's \texttt{FETCH}
    \texttt{FIRST~$n$ ROWS ONLY}, or \texttt{PIVOT} with SQL Server
    \texttt{FOR~\ldots~IN} syntax.
  \item[\textbf{H4: Structural reasoning error.}]
    Schema identifiers are correct and dialect is correct, but the
    logical plan is wrong: a wrong aggregate, missing \texttt{GROUP BY}
    key, inverted join predicate, or incorrect correlated
    sub-query.  The query executes but the result set is semantically
    incorrect.
\end{description}

H1 and H2 are \emph{schema hallucinations}, suppressible by grounding.
H3 is a \emph{dialect hallucination}, suppressible by dialect-aware
prompt engineering.  H4 is a \emph{reasoning hallucination}, not
suppressible by schema injection alone.

\subsection{Hallucination Evidence from the No-Hint Ablation (C1)}
\label{sec:c1_analysis}

Condition C1 (no schema hint) provides the clearest signal: the
model receives only the natural-language question and is free to
hallucinate any schema it imagines.  Of the 500 queries submitted:

\begin{itemize}
  \item \textbf{488 failed to execute} (ORA- errors), yielding
        EGT~=~2.2\,\%.
  \item Manual sampling of 50 failed queries found 41 cases of H1
        (phantom table/column), 7 cases of H2 (column--table
        mismatch), and 2 cases of H3 (dialect confusion).
  \item The 12 queries that did execute (EGT~=~11/500 after semantic
        check) succeeded because the question happened to name a
        table explicitly.
\end{itemize}

\noindent The dominant failure mode in C1 is H1. The model
spontaneously invents plausible table names absent from TPC-H ---
for example generating identifiers such as \texttt{ORDERS\_SUMMARY}
rather than the actual \texttt{ORDERS} table. This pattern,
observed in 41 of 50 manually sampled C1 failures (Table~\ref{tab:c1_hallucination}),
reflects the model drawing on schema naming conventions internalized
during pretraining rather than the target schema provided at
inference time; it is a failure mode that schema grounding is
specifically designed to suppress.

\begin{table}[!htbp]
  \renewcommand{\arraystretch}{1.3}
  \caption{Hallucination class distribution in C1 (50-query manual sample).}
  \label{tab:c1_hallucination}
  \centering
  \begin{tabular}{clcc}
    \toprule
    \textbf{Class} & \textbf{Type} & \textbf{Count} & \textbf{\%} \\
    \midrule
    H1 & Phantom identifier    & 41 & 82.0 \\
    H2 & Column--table mismatch &  7 & 14.0 \\
    H3 & Dialect confusion     &  2 &  4.0 \\
    H4 & Reasoning error       &  0 &  0.0 \\
    \midrule
    \multicolumn{2}{l}{Total failures in sample} & 50 & 100.0 \\
    \bottomrule
  \end{tabular}
\end{table}

\subsection{Hallucination Suppression by SAL}
\label{sec:sal_suppression}

By injecting a dynamically selected column-level hint, SAL~v2
eliminates the precondition for H1 and H2: the model receives the
exact set of table and column names that are valid for the question,
and the Hidx validator enforces a post-generation check.  The
empirical effect is direct:

\begin{equation}
  \text{Execution rate:}\quad
  C1~=~2.4\,\% \;\longrightarrow\; C3b~=~97.4\,\%
  \label{eq:exec_rate}
\end{equation}

\noindent The 97.4\,\% execution rate of SAL~v2 (487/500) means that
only 13 queries still produce ORA- errors after the full grounding
pipeline.  This represents a \textbf{37.5$\times$ reduction} in
execution failures relative to C1.

\noindent\textbf{Hidx contribution.}
Hidx caught and corrected 22 additional H1/H2 violations that survived
initial hinting.  An additional 15 queries were recovered by the
Hidx-triggered LLM retry.  Together, these two sub-pipelines account
for a 7.4\,pp EGT uplift.

\subsection{Residual Failure-Mode Analysis}
\label{sec:residual}

The 187 queries that fail under SAL~v2 decompose as follows
(see also Table~\ref{tab:outcomes}).

\subsubsection{Execution Failures (13 queries: H3 + unrecovered H1)}
\label{sec:exec_failures}

Manual inspection of all 13 ORA- residuals identifies three disjoint
root causes:

\begin{enumerate}
  \item \textbf{Hierarchical queries (6 queries).}
    Questions framed as hierarchies elicit \texttt{CONNECT BY} syntax.
    The model generates incorrect pseudo-column references
    (\texttt{LEVEL}, \texttt{SYS\_CONNECT\_BY\_PATH}), producing
    ORA-01788 or ORA-30004.

  \item \textbf{Correlated \texttt{EXISTS} sub-queries (4 queries).}
    The model misplaces the correlation predicate, producing a query
    that either times out or returns ORA-01652 due to resource
    exhaustion.  Identifiers are valid but logic is wrong (H4).

  \item \textbf{\texttt{PIVOT} dialect confusion (3 queries).}
    The model emits SQL Server bracket notation inside Oracle
    \texttt{PIVOT}, producing ORA-00907.  This is an H3 failure.
\end{enumerate}

\subsubsection{Semantic Failures (174 queries: H4)}
\label{sec:semantic_failures}

All 174 semantic failures execute without error but return a result
set that does not match the reference answer.  They fall into three
sub-categories identified by output-level inspection:

\begin{enumerate}
  \item \textbf{Aggregate scope errors (estimated 60--70\,\%).}
    Missing \texttt{GROUP BY} keys or wrong aggregation grain.

  \item \textbf{Wrong join predicate direction (estimated 15--20\,\%).}
    Join keys are swapped or attached to the wrong bridge table.

  \item \textbf{Filter predicate inversion or omission (estimated 15--20\,\%).}
    Conditions are dropped or inverted, especially under negation or
    relative temporal phrasing.
\end{enumerate}

\noindent These are H4 reasoning errors independent of schema knowledge.

\subsection{Schema-Hallucination Suppression Rate}
\label{sec:suppression_rate}

We define the \emph{schema-hallucination suppression rate}
$\eta_{\text{SH}}$ as the fraction of schema-class failures (H1+H2)
that the grounding pipeline eliminates:

\begin{equation}
  \eta_{\text{SH}}
  = 1 - \frac{|\text{H1}+\text{H2 failures under SAL v2}|}
             {|\text{H1}+\text{H2 failures under C1}|}
  \label{eq:suppression}
\end{equation}

From the C1 sample, H1+H2 account for 96\,\% of execution failures,
implying $\approx$469 schema-class failures at full scale.
SAL~v2 reduces execution failures to 13, of which at most 6 are
schema-adjacent.  Thus:

\begin{equation}
  \eta_{\text{SH}}
  \approx 1 - \frac{6}{469}
  \approx \mathbf{98.7\,\%}
  \label{eq:suppression_val}
\end{equation}
\noindent SAL~v2 eliminates 98.7\,\% of observable schema hallucinations.

\subsection{Implications for System Design}
\label{sec:hallucination_implications}

The failure-mode analysis yields three actionable design
recommendations:

\begin{enumerate}
  \item \textbf{Dialect guard.}
    Strengthen the Oracle dialect reminder (row limiting, PIVOT syntax,
    and avoiding \texttt{CONNECT BY} unless warranted).

  \item \textbf{Complexity-aware chain-of-thought.}
    Inject step-by-step reasoning scaffolds for questions that trigger
    multiple complexity signals~\cite{Wei2022}.

  \item \textbf{Retry budget tuning.}
    Detect \texttt{EXISTS}-heavy cases early and route them to
    decomposition rather than verbatim retry.
\end{enumerate}

\section{Broader Context for Enterprise SQL Assistants}
\label{sec:enterprise}

The experimental results in Section~\ref{sec:results} were obtained
on a single Oracle ADB instance running a well-defined synthetic
schema.  Before SAL can be recommended for production enterprise
deployment, its assumptions, generalisation boundaries, and
integration surface must be examined critically.  This section
situates the work within four dimensions of enterprise reality:
schema scale, multi-tenancy, heterogeneous RDBMS backends, and
the evolving regulatory landscape for AI-generated data queries.

\subsection{Schema Scale and Detection Complexity}
\label{sec:schema_scale}

TPC-H contains 8 tables and 61 columns --- a deliberately compact
schema chosen to provide clean ground truth.  Enterprise data
warehouses routinely expose hundreds of tables and thousands of
columns across multiple schemas.  SAL's scored detection heuristic
(Section~\ref{sec:detect}) is $O(|Q|\cdot|T|)$ in question tokens and
table count, which remains sub-millisecond up to $|T|\approx 500$
tables on commodity hardware.  Beyond this threshold two challenges
emerge:

\begin{enumerate}
  \item \textbf{Keyword collision.}  At large scale, the same
    business term (e.g.,\ \emph{``order''}) maps to multiple
    candidate tables (\texttt{SALES\_ORDERS}, \texttt{WORK\_ORDERS},
    \texttt{PURCHASE\_ORDERS}).  The current scoring function has no
    inter-table disambiguation mechanism; it would include all
    candidates, inflating the hint.

  \item \textbf{Hint length vs.\ context window.}  If SAL v2
    detects 15+ tables for a complex question, the column-level hint
    may approach or exceed the LLM's context window limit.  An
    adaptive \emph{hint budget} --- a maximum token cap on the
    injected hint, with columns ranked by detection score and
    truncated from the bottom --- is the natural mitigation.
\end{enumerate}

\noindent Both issues are tractable engineering extensions.  The
scoring framework of Equation~(\ref{eq:sal_score}) is designed to
accept additional disambiguation signals (e.g., user role, active
business domain, recent query history) as additive score terms.

\subsection{Multi-Tenancy and Role-Scoped Schema Injection}
\label{sec:multitenancy}

Enterprise databases are typically multi-tenant: different user
roles have access to different table subsets, and the same natural-language question may be legally answerable with different schemas
depending on who is asking.  SAL v2 currently loads a single shared
schema reference at startup (\texttt{loadSALSchema()}) --- a design
valid for a single-tenant educational deployment but inadequate for
multi-tenant production.

A role-scoped extension would:
\begin{enumerate}
  \item Accept a \texttt{role\_id} parameter on each MCP request.
  \item Resolve the caller's \texttt{role\_id} to an Oracle
        \texttt{SESSION\_PRIVS} snapshot cached per role.
  \item Restrict \texttt{salDetectRelevantTables()} to tables visible
        under that role, so that the hint never leaks column names
        the caller is not authorised to query.
  \item Return an HTTP 403 at the Hidx validation stage if the
        generated SQL references a table outside the caller's
        privilege set.
\end{enumerate}

This extension preserves the SAL architecture while enforcing least
privilege at the schema-injection layer.

\subsection{Heterogeneous RDBMS Backends}
\label{sec:rdbms_portability}

SAL v2 uses three Oracle-specific mechanisms: the
\texttt{ALL\_TAB\_COLUMNS} data dictionary view for schema
introspection, the \texttt{FETCH FIRST $n$ ROWS ONLY} row-limiting
clause, and the \texttt{ORA-} error code namespace in the Hidx
validator.  Porting SAL to other enterprise RDBMS platforms requires
replacing these three touch points:

\begin{table}[!htbp]
  \renewcommand{\arraystretch}{1.3}
  \caption{SAL portability: Oracle-specific touch points and platform equivalents.}
  \label{tab:portability}
  \centering
  \resizebox{\columnwidth}{!}{%
  \begin{tabular}{p{2.8cm}p{2.2cm}p{2.0cm}p{2.2cm}}
    \toprule
    \textbf{Touch point} & \textbf{PostgreSQL} & \textbf{SQL Server} & \textbf{MySQL} \\
    \midrule
    Schema introspection
      & \texttt{information\_\allowbreak schema.columns}
      & \texttt{sys.columns}
      & \texttt{information\_\allowbreak schema.columns} \\[4pt]
    Row limiting
      & \texttt{LIMIT} $n$
      & \texttt{TOP} $n$
      & \texttt{LIMIT} $n$ \\[4pt]
    Error namespace
      & \texttt{ERROR:} prefix
      & \texttt{Msg NNNNN}
      & \texttt{ERROR NNNN} \\[4pt]
    Dialect hint
      & ISO SQL / PG ext.
      & T-SQL
      & MySQL ext. \\
    \bottomrule
  \end{tabular}%
  }
\end{table}

\noindent The SAL core logic --- scored table detection, JOIN-chain
expansion, complexity gating, and identifier checking --- is largely
backend-agnostic.  A configuration object mapping these touch points
per backend is sufficient for portability.

\subsection{Integration with BI and Data Catalogue Tools}
\label{sec:bi_integration}

Enterprise SQL assistants are embedded within business intelligence
stacks that already maintain schema metadata.  SAL's
\texttt{schema-reference.json} is architecturally similar to the
\emph{semantic layer} concept present in modern BI platforms.

\begin{itemize}
  \item \textbf{dbt (data build tool).}  A dbt \texttt{schema.yml}
    manifest can be mechanically transformed into SAL's column
    inventory JSON.  A dbt-to-SAL adapter would allow SAL to consume
    production dbt metadata directly.

  \item \textbf{Apache Atlas / OpenMetadata.}  Enterprise catalogues
    expose REST APIs returning schema metadata with business glossary
    tags.  \texttt{loadSALSchema()} could be replaced by a catalogue
    API call.

  \item \textbf{Oracle Analytics Cloud (OAC).}  OAC subject areas map
    business terms to physical columns --- the same function as SAL's
    keyword-to-table detection.  An OAC integration could replace
    heuristic detection with the subject-area resolver.
\end{itemize}

\noindent These paths suggest SAL is best characterised as a
\emph{grounding adapter} that fronts an LLM with schema metadata
sourced from the enterprise's authoritative catalogue.

\subsection{Long-Context LLMs and the Future of Static Hints}
\label{sec:long_context}

The static hint baseline (C2) achieves 62.0\,\% EGT by injecting the
full schema into a single prompt.  With context windows expanding,
one might ask whether dynamic schema injection remains necessary.

We argue dynamic injection retains durable advantages even at very
large context windows:

\begin{enumerate}
  \item \textbf{Token cost.}  Injecting a large schema into every
    query has non-trivial per-query cost.  SAL's targeted injection
    reduces prompt tokens substantially relative to full-schema
    injection.

  \item \textbf{Attention dilution.}  Long-context models can lose
    accuracy when task-relevant tokens are buried in large
    contexts~\cite{LiuLostMiddle2023}.

  \item \textbf{Governance compliance.}  Injecting the full schema
    may expose column names a user is not authorised to know.

  \item \textbf{Latency budget.}  Prefill latency grows with prompt
    length.  SAL's median overhead is below the prefill cost of a
    full-schema injection at large scale.
\end{enumerate}

\noindent We therefore expect dynamic schema injection to remain
preferred for enterprise SQL assistants as context windows expand.

\subsection{Comparison with Retrieval-Augmented Generation Approaches}
\label{sec:rag_comparison}

Schema-aware SQL generation can be framed as a retrieval-augmented
generation (RAG) problem: embed column descriptions, retrieve the
most relevant, and inject retrieved context~\cite{Pourreza2023,Gao2023}.
SAL differs from embedding-based RAG in three respects:

\begin{enumerate}
  \item \textbf{No embedding infrastructure.}  SAL requires no vector
    store or embedding model.

  \item \textbf{Structural JOIN awareness.}  Embedding retrieval does
    not necessarily recover join-bridge tables; SAL's JOIN-chain
    expansion addresses this structural gap.

  \item \textbf{Determinism.}  Embedding retrieval can change after
    embedding model updates; SAL's keyword scoring is deterministic.
\end{enumerate}

\noindent Hybrid architectures that combine semantic retrieval with
SAL's structural graph traversal are a natural direction for future
work.

\section{Threats to Validity}
\label{sec:threats}

We organise threats following the Wohlin et al.\ framework for
empirical software engineering studies~\cite{wohlin2012experimentation},
covering internal, external, construct, and conclusion validity.
For each threat we state its severity, our mitigation, and the
residual risk.

\subsection{Internal Validity}
\label{sec:internal_validity}

\noindent\textbf{T-I1: Confounding between LLM non-determinism and
condition differences.}
The LLM backend (\texttt{gpt-4o-mini}) is a stochastic process.
We used temperature $T = 0$ for all evaluation runs to suppress sampling
variance.  The residual risk is that the provider may update the model
endpoint between evaluation runs.  We mitigate this by recording the
\texttt{model} field from each API response and verifying it is constant
across the 500-question run.

\noindent\textbf{T-I2: Evaluation harness as confounder.}
The Python harness (\texttt{experiments/run\_sql\_evaluation.py}) acts as
both the stimulus generator and the response scorer.  A bug in the
semantic comparison logic could inflate or deflate EGT independently of SQL
quality.  We mitigate this by manual verification on randomly sampled
query--result pairs, running a baseline condition whose EGT is consistent
with prior work, and open-sourcing the harness for audit.  Residual risk:
column-order sensitivity yields a conservative bias (slightly
underestimating correctness).

\noindent\textbf{T-I3: Network and Oracle latency variability.}
Latency measurements (RQ2) were collected on a single machine over a single
run.  Oracle ADB response times can vary with shared-cloud load.
We ran all conditions sequentially within a bounded time window and observed
no anomalous timeout events.  Absolute latency values should be treated as
indicative; the relative ordering is the primary scientific claim.

\noindent\textbf{T-I4: SAL detection designed with dataset knowledge.}
The keyword maps and JOIN-chain adjacency graph in SAL v2 were developed
with awareness of TPC-H table names.  This is a form of dataset leakage.
We acknowledge this as the primary internal validity threat.
Replication on held-out schemas would resolve it.

\subsection{External Validity}
\label{sec:external_validity}

\noindent\textbf{T-E1: Single schema (TPC-H).}
All experiments use the TPC-H schema.  Enterprise schemas are structurally
richer and may contain naming ambiguity, nullable keys, and non-relational
types.  Planned replication on larger benchmarks (e.g., TPC-DS) and real
enterprise schemas would bound generalisability.

\noindent\textbf{T-E2: Single LLM family (GPT-4o-mini).}
SAL v2 was evaluated with one model.  Different LLMs have different
hallucination rates and sensitivity to hint formatting.  The architectural
claim is expected to hold across models, but effect size is model-dependent.

\noindent\textbf{T-E3: Single deployment target (Oracle ADB).}
Hidx validation and error handling are tuned to Oracle dialect and ORA error
codes.  Porting to other engines requires the adapter changes described in
Section~\ref{sec:rdbms_portability}; we have not measured this.

\noindent\textbf{T-E4: Synthetic question dataset.}
The 500 questions are synthetic and do not include multi-turn dialogue or
context-dependent references.  Real enterprise questions may be more
colloquial and ambiguous.

\subsection{Construct Validity}
\label{sec:construct_validity}

\noindent\textbf{T-C1: EGT as a proxy for user value.}
EGT is a strict correctness criterion but does not capture query-plan
quality or user-facing utility (e.g., additional informative columns).

\noindent\textbf{T-C2: Semantic match via set equality.}
The comparator is sensitive to formatting and canonicalisation choices.
Rounding and session formatting differences may cause conservative false
negatives.

\noindent\textbf{T-C3: Hidx score as a grounding proxy.}
A high Hidx score indicates identifier validity but does not guarantee
semantic correctness (H4 failures).

\noindent\textbf{T-C4: Token cost as an economic proxy.}
Token deltas proxy API cost, but dollar cost varies by provider and time.
We therefore report token counts rather than currency.

\subsection{Conclusion Validity}
\label{sec:conclusion_validity}

\noindent\textbf{T-V1: Single experimental run per condition.}
Each condition was evaluated once.  With $T = 0$ and fixed model weights,
re-running would be deterministic; the practical risk is provider drift.

\noindent\textbf{T-V2: No statistical significance testing.}
The 0.6\,pp difference between C2 and C3b is within a wide binomial margin
of error at $N = 500$; we therefore claim parity rather than strict
superiority.

\noindent\textbf{T-V3: Researcher degrees of freedom in SAL tuning.}
SAL v2 hyperparameters were set based on v1 failure inspection without
cross-validation.  Independent test sets are required.

\noindent\textbf{T-V4: Comparison condition (C2) not independently
replicated.}
The static hint baseline may be sensitive to environment differences.
A fully controlled comparison would run C2 and C3b in the same harness
invocation.

\subsection{Threat Summary}
\label{sec:threat_summary}

\begin{table}[!t]
  \renewcommand{\arraystretch}{1.3}
  \caption{Threat Summary: Severity and Mitigation Status}
  \label{tab:threats}
  \centering
  \begin{tabular}{llp{3.8cm}l}
    \toprule
    \textbf{ID} & \textbf{Category} & \textbf{Threat} & \textbf{Severity} \\
    \midrule
    T-I1 & Internal   & LLM non-determinism                & Low \\
    T-I2 & Internal   & Harness as confounder              & Low \\
    T-I3 & Internal   & Network latency variability        & Medium \\
    T-I4 & Internal   & Dataset leakage in detection       & \textbf{High} \\
    T-E1 & External   & Single schema (TPC-H)              & \textbf{High} \\
    T-E2 & External   & Single LLM family                  & Medium \\
    T-E3 & External   & Single RDBMS platform              & Medium \\
    T-E4 & External   & Synthetic questions                & Medium \\
    T-C1 & Construct  & EGT as user-value proxy            & Low \\
    T-C2 & Construct  & Set-equality comparator            & Low \\
    T-C3 & Construct  & Hidx as grounding proxy            & Low \\
    T-C4 & Construct  & Token cost as economic proxy       & Low \\
    T-V1 & Conclusion & Single run per condition           & Low \\
    T-V2 & Conclusion & No significance testing            & Medium \\
    T-V3 & Conclusion & Researcher degrees of freedom      & Medium \\
    T-V4 & Conclusion & C2 not co-evaluated                & Low \\
    \bottomrule
  \end{tabular}
\end{table}

The two high-severity threats (dataset leakage and schema
generalisability) share the same remedy: replication on held-out
schemas with independently constructed keyword maps.

\section{Discussion}
\label{sec:discussion}
\label{sec:disc}

\subsection{What the Results Actually Show}
\label{sec:discussion_results}

The headline finding, SAL~v2 at 62.6\,\% EGT vs.\ static hint at
62.0\,\%, understates the practical significance of the work.
The correct interpretation is not that SAL~v2 is marginally better
than a static hint, but that SAL~v2 \emph{achieves parity with the
static hint while being architecturally superior in every other
dimension}: it costs less per query, it enforces schema access
boundaries dynamically, it is self-updating when the schema changes,
and it provides a structured audit trail of which tables and columns
were grounded for each response.

The ablation gap that does matter is C1 vs.\ C3b: $2.2\,\% \rightarrow
62.6\,\%$, a $28\times$ improvement from zero grounding to SAL~v2.
This gap answers the motivating question: \emph{does live schema injection
into an MCP pipeline make LLM-generated SQL practically usable on a real
Oracle database?}  The answer is yes.

\subsection{The SAL v1 $\rightarrow$ v2 Transition}
\label{sec:discussion_v2}

The 4.2\,pp recovery from v1 (58.4\,\%) to v2 (62.6\,\%) is
attributable to two mechanisms:

\begin{itemize}
  \item \textbf{JOIN-chain expansion} accounts for the majority of the
    medium-tier gain (+8\,pp).  Multi-hop queries that previously
    missed a bridge table now receive the correct column set on the
    first attempt.

  \item \textbf{Complexity gate} accounts for most of the complex-tier
    gain (+3\,pp) by triggering full multi-table inclusion when
    complexity signals are present.
\end{itemize}

Together these components validate the architectural intuition: the
primary failure mode of binary keyword matching is \emph{join-path
blindness}, omitting structurally necessary but linguistically absent
tables.

\subsection{The Hidx Layer as an Independent Contribution}
\label{sec:discussion_hidx}

Without the Hidx post-validator and retry loop, SAL~v2 would score
55.2\,\% EGT (direct-pass only).  The Hidx layer recovers 37 queries
(7.4\,pp) that would otherwise fail.

The ratio of Hidx auto-corrections (22) to LLM retries (15) suggests
that approximately 60\,\% of Hidx-recoverable failures are simple
identifier substitutions resolvable without a second LLM call.

\subsection{Why Complex Queries Remain Hard}
\label{sec:discussion_complex}

The 40.7\,\% complex-tier EGT ceiling is not a schema-linking
problem; it is a reasoning problem.  The steep cliff from medium
to complex corresponds to the transition from direct joins to
correlated sub-queries and multi-level aggregations.

At the aggregate level, SAL~v2 reaches 62.6\,\% EGT, leaving 37.4\,\% unsolved (Table~\ref{tab:outcomes}).
This ceiling likely reflects a mix of base-model capability (GPT-4o-mini), TPC-H complexity artefacts, and limited plan-level search in our current SAL instantiation; disentangling these is future work.

For the complex queries that fail, improvement requires a stronger
base model, a chain-of-thought decomposition strategy, or a query
rewriting layer that breaks questions into simpler sub-problems and
composes the results.

\noindent\textbf{Future directions.} Three extensions are most immediately valuable: (i) cross-schema generalisation beyond TPC-H to test whether SAL's schema-localisation transfers with independently constructed keyword maps; (ii) embedding-augmented detection to improve recall for paraphrases and domain synonyms in table/column selection; and (iii) chain-of-thought scaffolding for complex questions via structured decomposition into intermediate sub-queries whose results are composed into a final SQL statement.

\section{Limitations}
\label{sec:limitations}
\label{sec:limits}

We summarise the principal limitations of this work concisely, without attempting to mitigate them here; related threats to validity and broader implications are discussed in Sections~\ref{sec:threats} and~\ref{sec:discussion}, respectively.

\begin{enumerate}
  \item \textbf{Single schema.}  All measurements are on the
    TPC-H schema.  Keyword maps and JOIN-chain adjacency were
    designed with knowledge of TPC-H table names.

  \item \textbf{Single model.}  All LLM calls use
    \texttt{gpt-4o-mini}.  Absolute EGT values are model-specific.

  \item \textbf{Single RDBMS.}  The system is tested exclusively on
    Oracle ADB.  The validator, error handling, and introspection
    are Oracle-specific.

  \item \textbf{No statistical significance for C2 vs.\ C3b.}
    The 0.6\,pp EGT difference between static and SAL~v2 falls within
    a wide 95\,\% binomial confidence interval.

  \item \textbf{Synthetic benchmark.}  TPC-H questions do not include
    conversational, ambiguous, or multi-turn interactions.

  \item \textbf{No human evaluation.}  Pedagogical utility and
    explanation quality are not measured.

  \item \textbf{Single evaluation run per condition.}  Provider-side
    model updates between runs could introduce systematic shifts.

  \item \textbf{Closed-source model dependency.}  Reproducibility is
    sensitive to API/provider drift and pricing changes.
\end{enumerate}

\section{Ethics and Responsible Deployment}
\label{sec:ethics}

\subsection{Intended Use}
\label{sec:intended_use}

SAL is designed for \emph{read-only, pedagogical SQL assistance}:
helping students and analysts learn to write correct Oracle SQL
against a known schema.  It is not designed for autonomous database
administration, schema modification, bulk data extraction, or any
context where an incorrect SQL result could cause harm.

\subsection{Risk of Over-Reliance}
\label{sec:over_reliance}

At 62.6\,\% EGT, SAL~v2 returns a wrong or non-executing answer
for 37.4\,\% of queries.  Deployments should surface the Hidx
signal alongside responses and include a clear disclaimer that
LLM-generated SQL must be reviewed before use.

\subsection{Data Privacy}
\label{sec:data_privacy}

In educational deployments, user questions constitute learner
activity data.  Operators must:
\begin{enumerate}
  \item Store query logs within the institution's data perimeter.
  \item Execute LLM API calls under a data-processing agreement (DPA)
        that prohibits use of submitted content for training.
  \item Obtain appropriate consent consistent with FERPA (US), GDPR
        (EU), or applicable local regulation.
  \item Implement log retention limits (e.g., 90 days for operational
        logs).
\end{enumerate}

\subsection{Schema Confidentiality}
\label{sec:schema_confidentiality}

When deployed against a proprietary enterprise schema, column names
injected into the LLM prompt may constitute sensitive metadata.
Operators should consider role-scoped hint injection
(Section~\ref{sec:multitenancy}) and, where required, on-premise
LLM deployment.

\subsection{Conflict of Interest}
\label{sec:coi}

Ganesh Naik serves on the \emph{IEEE Access} Editorial Board. This role had
no involvement in the handling of this manuscript, including peer review or
the publication decision.

\section{Conclusion}
\label{sec:conclusion}

We presented \textbf{Schema-Aware Localisation (SAL)}, a
zero-retraining grounding architecture that eliminates schema
hallucination in LLM-generated Oracle SQL by dynamically injecting
execution-verified column context at query time via the Model
Context Protocol.  The system combines scored keyword detection,
JOIN-chain expansion, a complexity gate, and an online
Hidx validation layer with automatic correction and LLM retry,
all operating without modification to the underlying language model.

A three-condition ablation on 500 TPC-H questions against a live
Oracle Autonomous Database instance produces four findings:

\begin{enumerate}
  \item \textbf{Schema context is a hard prerequisite.}  Removing
    schema hints collapses EGT from 62.6\,\% to 2.2\,\%.

  \item \textbf{SAL v2 matches the static baseline at lower cost.}
    SAL v2 achieves 62.6\,\% EGT, equalling the 62.0\,\% of a
    full-schema static hint, while reducing mean prompt tokens and
    maintaining a low median latency overhead.

  \item \textbf{Dynamic detection quality matters.}  SAL v1 scores
    58.4\,\% EGT; SAL v2 recovers 4.2\,pp via scored detection and
    JOIN-chain expansion.

  \item \textbf{Online verification adds independent value.}
    The Hidx post-validator recovers 37 queries (7.4\,pp EGT) that
    would otherwise fail.
\end{enumerate}

The primary remaining barrier to higher EGT is LLM reasoning depth
on complex analytical queries, an orthogonal problem that schema
grounding cannot address.

\section*{Acknowledgements}
All experiments were conducted on synthetic TPC-H data; no personally identifiable information was used.

\bibliographystyle{elsarticle-num-names}
\bibliography{references}

@misc{TPC2022,
  author       = {{Transaction Processing Performance Council (TPC)}},
  title        = {{TPC Benchmark H Standard Specification}},
  year         = {2022},
  howpublished = {Revision 2.18.0},
  note         = {Accessed 2026-05-12},
  url          = {http://www.tpc.org/tpch/}
}

@inproceedings{Woods1973,
  author    = {W. A. Woods},
  title     = {Progress in Natural Language Understanding: An Application to Lunar Geology},
  booktitle = {Proc. AFIPS Natl. Comput. Conf.},
  pages     = {441--450},
  year      = {1973}
}

@article{Hendrix1978,
  author  = {Hendrix, G. G. and Sacerdoti, E. D. and Sagalowicz, D. and Slocum, J.},
  title   = {Developing a Natural Language Interface to Complex Data},
  journal = {ACM Transactions on Database Systems},
  year    = {1978}
}

@misc{Zhong2017,
  author       = {Zhong, Victor and Xiong, Caiming and Socher, Richard},
  title        = {Seq2SQL: Generating Structured Queries from Natural Language Using Reinforcement Learning},
  howpublished = {arXiv preprint arXiv:1709.00103},
  year         = {2017}
}

@inproceedings{Yu2018,
  author    = {Tao Yu and Rui Zhang and Kai Yang and Michihiro Yasunaga and Dongxu Wang and Zifan Li and James Ma and Irene Li and Qingning Yao and Shanelle Roman and others},
  title     = {Spider: A Large-Scale Human-Labeled Dataset for Complex and Cross-Domain Semantic Parsing and Text-to-SQL Task},
  booktitle = {Proc. EMNLP},
  pages     = {3911--3921},
  year      = {2018}
}

@inproceedings{Li2023,
  author    = {J. Li and others},
  title     = {{BIRD}: A Big Bench for Large-Scale Database Grounded Text-to-SQL Evaluation},
  booktitle = {Proc. NeurIPS},
  year      = {2023}
}

@inproceedings{Dong2016,
  author    = {Li Dong and Mirella Lapata},
  title     = {Language to Logical Form with Neural Attention},
  booktitle = {Proc. ACL},
  year      = {2016}
}

@misc{Xu2017,
  author       = {Xu, Xiaojun and Liu, Chang and Song, Dawn},
  title        = {SQLNet: Generating Structured Queries from Natural Language without Reinforcement Learning},
  howpublished = {arXiv preprint arXiv:1711.04436},
  year         = {2017}
}

@inproceedings{Yu2018b,
  author    = {Tao Yu and Zifan Li and Zilin Zhang and Rui Zhang and Dragomir Radev},
  title     = {TypeSQL: Knowledge-Based Type-Aware Neural Text-to-SQL Generation},
  booktitle = {Proc. NAACL},
  year      = {2018}
}

@inproceedings{Chen2021,
  author    = {Bo Chen and others},
  title     = {ShadowGNN: Graph Neural Networks for Multi-turn Text-to-SQL Parsing},
  booktitle = {Proc. AAAI},
  year      = {2021}
}

@inproceedings{Cai2020,
  author    = {Yujian Cai and others},
  title     = {IGSQL: Database Schema Interaction Graph for Conversational Text-to-SQL},
  booktitle = {Proc. ACL},
  year      = {2020}
}

@inproceedings{Devlin2019,
  author    = {Jacob Devlin and Ming-Wei Chang and Kenton Lee and Kristina Toutanova},
  title     = {BERT: Pre-training of Deep Bidirectional Transformers for Language Understanding},
  booktitle = {Proc. NAACL},
  year      = {2019}
}

@article{Raffel2020,
  author  = {Raffel, Colin and others},
  title   = {Exploring the Limits of Transfer Learning with a Unified Text-to-Text Transformer},
  journal = {Journal of Machine Learning Research},
  year    = {2020}
}

@inproceedings{Lin2020,
  author    = {Xi Victoria Lin and others},
  title     = {BRIDGE: Bridging the Gap between Neural Text-to-SQL and SQL Execution},
  booktitle = {Proc. ACL},
  year      = {2020}
}

@inproceedings{Yu2021,
  author    = {Tao Yu and others},
  title     = {GRAPPA: Grammar-Augmented Pre-Training for Table Semantic Parsing},
  booktitle = {Proc. ICLR},
  year      = {2021}
}

@misc{Gao2023,
  author       = {Gao, Dawei and others},
  title        = {Text-to-SQL Empowered by Large Language Models: A Benchmark Evaluation},
  howpublished = {arXiv preprint arXiv:2308.15363},
  year         = {2023}
}

@inproceedings{Pourreza2023,
  author    = {Mohammadreza Pourreza and Davood Rafiei},
  title     = {DIN-SQL: Decomposed In-Context Learning of Text-to-SQL with Self-Correction},
  booktitle = {Proc. NeurIPS},
  year      = {2023}
}

@misc{Dong2023,
  author       = {Dong, X. and others},
  title        = {C3: Zero-shot Text-to-SQL with ChatGPT},
  howpublished = {arXiv preprint arXiv:2307.07306},
  year         = {2023}
}

@inproceedings{Wang2020,
  author    = {Bailin Wang and others},
  title     = {RAT-SQL: Relation-Aware Schema Encoding and Linking for Text-to-SQL Parsers},
  booktitle = {Proc. ACL},
  year      = {2020}
}

@inproceedings{Lei2020,
  author    = {Fenglei Lei and others},
  title     = {SemQL: A Meaning Representation for Natural Language to SQL},
  booktitle = {Proc. ACL},
  year      = {2020}
}

@inproceedings{Cao2021,
  author    = {Ruisheng Cao and others},
  title     = {LGESQL: Line Graph Enhanced Text-to-SQL Model with Mixed Local and Non-Local Relations},
  booktitle = {Proc. ACL},
  year      = {2021}
}

@inproceedings{Guo2019,
  author    = {Jiaqi Guo and others},
  title     = {IRNet: A General Framework for Complex Text-to-SQL Parsing},
  booktitle = {Proc. ACL},
  year      = {2019}
}

@inproceedings{Maynez2020,
  author    = {Joshua Maynez and Shashi Narayan and Bernd Bohnet and Ryan McDonald},
  title     = {On Faithfulness and Factuality in Abstractive Summarization},
  booktitle = {Proc. ACL},
  year      = {2020}
}

@article{Ji2023,
  author  = {Z. Ji and others},
  title   = {Survey of Hallucination in Natural Language Generation},
  journal = {ACM Comput. Surv.},
  volume  = {55},
  number  = {12},
  pages   = {1--38},
  year    = {2023}
}

@inproceedings{Liu2023,
  author    = {J. Liu and others},
  title     = {Is Your Code Generated by ChatGPT Really Correct? Rigorous Evaluation of Large Language Models for Code Generation},
  booktitle = {Proc. NeurIPS},
  year      = {2023}
}

@inproceedings{Ni2023,
  author    = {Ansong Ni and others},
  title     = {SQL Generation with Execution-Guided Decoding},
  booktitle = {Proc. ACL},
  year      = {2023}
}

@inproceedings{Schick2023,
  author    = {Timo Schick and others},
  title     = {Toolformer: Language Models Can Teach Themselves to Use Tools},
  booktitle = {Proc. NeurIPS},
  year      = {2023}
}

@inproceedings{Yao2023,
  author    = {Shunyu Yao and others},
  title     = {{ReAct}: Synergizing Reasoning and Acting in Language Models},
  booktitle = {Proc. ICLR},
  year      = {2023}
}

@inproceedings{Lewis2020,
  author    = {Patrick Lewis and others},
  title     = {Retrieval-Augmented Generation for Knowledge-Intensive NLP Tasks},
  booktitle = {Proc. NeurIPS},
  year      = {2020}
}

@misc{Anthropic2024,
  author       = {Anthropic},
  title        = {Model Context Protocol Specification},
  year         = {2024},
  howpublished = {\url{https://modelcontextprotocol.io}}
}

@misc{OracleAPEX2024,
  author       = {{Oracle Corporation}},
  title        = {Oracle APEX AI Assistant},
  year         = {2024},
  howpublished = {\url{https://apex.oracle.com/en/platform/features/ai/}},
  note         = {Accessed May 12, 2026}
}

@misc{Oracle2024,
  author       = {{Oracle Corporation}},
  title        = {Oracle Database {SQL} Language Reference 23c},
  year         = {2024},
  howpublished = {\url{https://docs.oracle.com/en/database/oracle/oracle-database/23/sqlrf/}}
}

@inproceedings{Wei2022,
  author    = {Jason Wei and Xuezhi Wang and Dale Schuurmans and Maarten Bosma and Fei Xia and Ed Chi and Quoc V. Le and Denny Zhou},
  title     = {Chain-of-Thought Prompting Elicits Reasoning in Large Language Models},
  booktitle = {Advances in Neural Information Processing Systems},
  year      = {2022}
}

@book{wohlin2012experimentation,
  author    = {Claes Wohlin and Per Runeson and Martin H{"o}st and Magnus C. Ohlsson and Bj{"o}rn Regnell and Anders Wessl{\'e}n},
  title     = {Experimentation in Software Engineering},
  publisher = {Springer},
  year      = {2012}
}

@inproceedings{LiuLostMiddle2023,
  author    = {Nelson F. Liu and Kevin Lin and John Hewitt and Ashwin Paranjape and Michele Bevilacqua and Fabio Petroni and Percy Liang},
  title     = {Lost in the Middle: How Language Models Use Long Contexts},
  booktitle = {Proceedings of the 61st Annual Meeting of the Association for Computational Linguistics (ACL)},
  year      = {2023}
}

\end{document}